\documentclass[manuscript,sigconf]{acmart}

\AtBeginDocument{%
  \providecommand\BibTeX{{%
    \normalfont B\kern-0.5em{\scshape i\kern-0.25em b}\kern-0.8em\TeX}}}

\copyrightyear{2022}
\acmYear{2022}
\setcopyright{rightsretained} 
\acmConference[MMSports '22]{Proceedings of the 5th International ACM
Workshop on Multimedia Content Analysis in Sports}{October 10,
2022}{Lisboa, Portugal}
\acmBooktitle{Proceedings of the 5th International ACM Workshop on
Multimedia Content Analysis in Sports (MMSports '22), October 10, 2022,
Lisboa, Portugal}\acmDOI{10.1145/3552437.3555703}
\acmISBN{978-1-4503-9488-8/22/10}

\usepackage{etoolbox}
\makeatletter
\patchcmd{\maketitle}{\@copyrightpermission}{
   \begin{minipage}{0.3\columnwidth}
     \href{https://creativecommons.org/licenses/by/4.0/}{\includegraphics[width=0.90\textwidth]{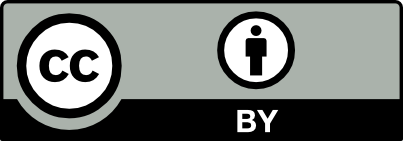}}
   \end{minipage}\hfill
   \begin{minipage}{0.7\columnwidth}
     \href{https://creativecommons.org/licenses/by/4.0/}{This work is licensed under a Creative Commons Attribution International 4.0 License.}
   \end{minipage}
  
   \vspace{5pt}
}{}{}

\makeatother

\acmSubmissionID{mmspor05}

\usepackage{multirow}
\newcommand{\mysection}[1]{\vspace{2mm}\noindent\textbf{#1.}}
\usepackage[symbol]{footmisc}
\usepackage{xspace}
\newcommand{\etal}{\emph{et al.}\xspace}
\newcommand{\eg}{\emph{e.g.}\xspace}
\newcommand{\ie}{\emph{i.e.}\xspace}
\usepackage{balance}
\begin{document}

\title{SoccerNet 2022 Challenges Results}

\newcommand{\tsc}[1]{\textsuperscript{#1}} 

\setlength{\emergencystretch}{1em}

\author{Silvio Giancola} 
\authornote{Both authors contributed equally to this research.}
\authornote{SoccerNet Organizers.}
\email{silvio.giancola@kaust.edu.sa}
\affiliation{  
	\institution{KAUST} 
 	\city{Thuwal} 
	\country{ Saudi Arabia} 
 }
\author{Anthony Cioppa}
\authornotemark[1]
\authornotemark[2]
\email{anthony.cioppa@uliege.be}
\affiliation{  
	\institution{University of Liège} 
 	\city{Liège} 
	\country{ Belgium} 
 }
\author{Adrien Deliège}
\authornotemark[2]
\email{adrien.deliege@uliege.be}
\affiliation{  
	\institution{University of Liège} 
 	\city{Liège} 
	\country{ Belgium} 
 }
\author{Floriane Magera}
\authornotemark[2]
\email{f.magera@evs.com}
\affiliation{  
	\institution{University of Liège, \& EVS Broadcast Equipment} 
 	\city{Liège} 
	\country{ Belgium} 
 }
\author{Vladimir Somers} 
\authornotemark[2]
\email{v.somers@sportradar.com}
\affiliation{  
	\institution{Sportradar, \& UCLouvain, \& EPFL} 
 	\city{London} 
	\country{ United Kingdom} 
 }
\author{Le Kang}
\authornotemark[2]
\email{deepconv@gmail.com }
\affiliation{  
	\institution{Baidu Research} 
 	\city{Sunnyvale} 
	\country{ USA} 
 }
\author{Xin Zhou} 
\authornotemark[2]
\email{chow459@gmail.com }
\affiliation{  
	\institution{Baidu Research} 
 	\city{Sunnyvale} 
	\country{ USA} 
 }
\author{Olivier Barnich} 
\authornotemark[2]
\email{o.barnich@evs.com}
\affiliation{  
	\institution{EVS Broadcast Equipment} 
 	\city{Liège} 
	\country{ Belgium} 
 }
\author{Christophe De Vleeschouwer}
\authornotemark[2]
\email{christophe.devleeschouwer@uclouvain.be}
\affiliation{  
	\institution{UCLouvain} 
 	\city{Louvain-la-Neuve} 
	\country{ Belgium} 
 }
\author{Alexandre Alahi}
\authornotemark[2]
\email{alexandre.alahi@epfl.ch}
\affiliation{  
	\institution{EPFL} 
 	\city{Lausanne} 
	\country{ Switzerland} 
 }
\author{Bernard Ghanem}
\authornotemark[2]
\email{bernard.ghanem@kaust.edu.sa}
\affiliation{  
	\institution{KAUST} 
 	\city{Thuwal} 
	\country{ Saudi Arabia} 
 }
\author{Marc Van Droogenbroeck}
\authornotemark[2]
\email{M.VanDroogenbroeck@uliege.be}
\affiliation{  
	\institution{University of Liège} 
 	\city{Liège} 
	\country{ Belgium} 
 }
\author{Abdulrahman Darwish} 
\email{abdulrahman.darwish@guc.edu.eg}
\affiliation{  
	\institution{German University in Cairo} 
 	\city{New Cairo City} 
	\country{ Egypt} 
 }
\author{Adrien Maglo} 
\email{adrien.maglo@cea.fr}
\affiliation{  
	\institution{Université Paris-Saclay,  CEA,  List} 
 	\city{Paris} 
	\country{ France} 
 }
\author{Albert Clapés} 
\email{alcl@create.aau.dk}
\affiliation{  
	\institution{Aalborg University} 
 	\city{Aalborg} 
	\country{ Denmark} 
 }
\author{Andreas Luyts} 
\email{andreas.luyts@rebatch.be}
\affiliation{  
	\institution{ReBatch} 
 	\city{Kontich} 
	\country{ Belgium} 
 }
\author{Andrei Boiarov} 
\email{andrei.boiarov@sit.team}
\affiliation{  
	\institution{Schaffhausen Institute of Technology} 
 	\city{Schaffhausen } 
	\country{ Switzerland} 
 }
\author{Artur Xarles} 
\email{arturxe@gmail.com}
\affiliation{  
	\institution{Universitat de Barcelona} 
 	\city{Barcelona} 
	\country{ Spain} 
 }
\author{Astrid Orcesi} 
\email{astrid.orcesi@cea.fr}
\affiliation{  
	\institution{Université Paris-Saclay,  CEA,  List} 
 	\city{Paris} 
	\country{ France} 
 }
\author{Avijit Shah} 
\email{avijit.shah@yahooinc.com}
\affiliation{  
	\institution{Yahoo Research} 
 	\city{Sunnyvale} 
	\country{ USA} 
 }
\author{Baoyu Fan} 
\email{fanbaoyu@inspur.com}
\affiliation{  
	\institution{Inspur Electronic Information Industry Co., Ltd. State Key Laboratory of High-end Server Storage Technology} 
 	\city{Jinan} 
	\country{ China} 
 }
\author{Bharath Comandur} 
\email{cjrbharath@gmail.com}
\affiliation{  
	\institution{Purdue University} 
 	\city{West Lafayette} 
	\country{USA} 
 }
\author{Chen Chen} 
\email{chenchen@oppo.com}
\affiliation{  
	\institution{OPPO Research Institute} 
 	\city{Shenzhen } 
	\country{ China} 
 }
\author{Chen Zhang} 
\email{zhangchen4@oppo.com}
\affiliation{  
	\institution{OPPO Research Institute} 
 	\city{Shenzhen } 
	\country{ China} 
 }
\author{Chen Zhao} 
\email{zhaochen03@baidu.com}
\affiliation{  
	\institution{Department of Augmented Reality Technology (ART),  Baidu Inc.} 
 	\city{Beijing} 
	\country{ China} 
 }
\author{Chengzhi Lin} 
\email{linchzh3@mail2.sysu.edu.cn}
\affiliation{  
	\institution{Sun Yat-sen University} 
 	\city{Guangzhou} 
	\country{ China} 
 }
\author{Cheuk-Yiu Chan} 
\email{cy3chan@cihe.edu.hk}
\affiliation{  
	\institution{Caritas Institute of Higher Education} 
 	\city{Tseung Kwan O} 
	\country{ Hong Kong,  SAR} 
 }
\author{Chun-Chuen Hui} 
\email{cchui@cihe.edu.hk}
\affiliation{  
	\institution{Caritas Institute of Higher Education} 
 	\city{Tseung Kwan O} 
	\country{ Hong Kong,  SAR} 
 }
\author{Dengjie Li} 
\email{lidengjie@meituan.com}
\affiliation{  
	\institution{Meituan Inc.} 
 	\city{Beijing} 
	\country{ China} 
 }
\author{Fan Yang} 
\email{fan.yang@fujitsu.com}
\affiliation{  
	\institution{Fujitsu Research} 
 	\city{Kawasaki} 
	\country{ Japan} 
 }
\author{Fan Liang} 
\email{liangfan02@meituan.com}
\affiliation{  
	\institution{Meituan Inc.} 
 	\city{Beijing} 
	\country{ China} 
 }
\author{Fang Da} 
\email{fang@qcraft.ai}
\affiliation{  
	\institution{QCraft Inc.} 
 	\city{Beijing} 
	\country{ China} 
 }
\author{Feng Yan} 
\email{yanfeng05@meituan.com}
\affiliation{  
	\institution{Meituan Inc.} 
 	\city{Beijing} 
	\country{ China} 
 }
\author{Fufu  Yu} 
\email{fufuyu@tencent.com}
\affiliation{  
	\institution{Tencent Youtu Lab} 
 	\city{Shanghai} 
	\country{ China} 
 }
\author{Guanshuo Wang} 
\email{mediswang@tencent.com}
\affiliation{  
	\institution{Tencent Youtu Lab} 
 	\city{Shanghai} 
	\country{ China} 
 }
\author{H. Anthony Chan} 
\email{hhchan@cihe.edu.hk}
\affiliation{  
	\institution{Caritas Institute of Higher Education} 
 	\city{Tseung Kwan O} 
	\country{ Hong Kong,  SAR} 
 }
\author{He Zhu} 
\email{zhuh20@mails.tsinghua.edu.cn}
\affiliation{  
	\institution{Tsinghua University} 
 	\city{Beijing} 
	\country{ China} 
 }
\author{Hongwei Kan} 
\email{kanhongwei@inspur.com}
\affiliation{  
	\institution{Inspur Electronic Information Industry Co., Ltd. State Key Laboratory of High-end Server Storage Technology} 
 	\city{Jinan} 
	\country{ China} 
 }
\author{Jiaming Chu} 
\email{chujiaming886@bupt.edu.cn}
\affiliation{  
	\institution{OPPO Research Institute, \& Beijing University of Posts and Telecommunications} 
 	\city{Shenzhen } 
	\country{ China} 
 }
\author{Jianming Hu} 
\email{hujm@mail.tsinghua.edu.cn}
\affiliation{  
	\institution{Tsinghua University} 
 	\city{Beijing} 
	\country{ China} 
 }
\author{Jianyang Gu} 
\email{gu_jianyang@zju.edu.cn}
\affiliation{  
	\institution{OPPO Research Institute, \& Zhejiang University} 
 	\city{Shenzhen } 
	\country{ China} 
 }
\author{Jin Chen} 
\email{chenjing@mgtv.com}
\affiliation{  
	\institution{MGTV} 
 	\city{Changsha} 
	\country{ China} 
 }
\author{João V. B. Soares} 
\email{jvbsoares@yahooinc.com}
\affiliation{  
	\institution{Yahoo Research} 
 	\city{Sunnyvale} 
	\country{ USA} 
 }
\author{Jonas Theiner} 
\email{theiner@l3s.de}
\affiliation{  
	\institution{L3S Research Center,  Leibniz University Hannover} 
 	\city{Hannover} 
	\country{ Germany} 
 }
\author{Jorge De Corte} 
\email{jorge.decorte@rebatch.be}
\affiliation{  
	\institution{ReBatch} 
 	\city{Kontich} 
	\country{ Belgium} 
 }
\author{José Henrique Brito} 
\email{jbrito@ipca.pt}
\affiliation{  
	\institution{2Ai – School of Technology  IPCA} 
 	\city{São Martinho} 
	\country{ Portugal} 
 }
\author{Jun Zhang} 
\email{bobbyjzhang@tencent.com}
\affiliation{  
	\institution{Tencent Youtu Lab} 
 	\city{Shanghai} 
	\country{ China} 
 }
\author{Junjie  Li} 
\email{serenitycapo@gmail.com}
\affiliation{  
	\institution{Tencent Youtu Lab, \& Shanghai Jiao Tong University} 
 	\city{Shanghai} 
	\country{ China} 
 }
\author{Junwei Liang} 
\email{junweiliang1114@gmail.com}
\affiliation{  
	\institution{Tencent Youtu Lab} 
 	\city{Shanghai} 
	\country{ China} 
 }
\author{Leqi Shen} 
\email{lunarshen@gmail.com}
\affiliation{  
	\institution{Tsinghua University} 
 	\city{Beijing} 
	\country{ China} 
 }
\author{Lin Ma} 
\email{linma@alumni.cuhk.net}
\affiliation{  
	\institution{Meituan Inc.} 
 	\city{Beijing} 
	\country{ China} 
 }
\author{Lingchi Chen} 
\email{lingchi@mgtv.com}
\affiliation{  
	\institution{MGTV} 
 	\city{Changsha} 
	\country{ China} 
 }
\author{Miguel Santos Marques} 
\email{a18888@alunos.ipca.pt}
\affiliation{  
	\institution{2Ai – School of Technology  IPCA} 
 	\city{São Martinho} 
	\country{ Portugal} 
 }
\author{Mike Azatov} 
\email{mazatov@gmail.com }
\affiliation{  
	\institution{Arsenal FC} 
 	\city{London} 
	\country{ United Kingdom} 
 }
\author{Nikita Kasatkin} 
\email{nk@sit.team}
\affiliation{  
	\institution{Schaffhausen Institute of Technology} 
 	\city{Schaffhausen}
	\country{Switzerland}
 }
\author{Ning Wang} 
\email{wangning12@mail.ecust.edu.cn}
\affiliation{  
	\institution{OPPO Research Institute, \& East China University of Science and Technology} 
 	\city{Shenzhen } 
	\country{ China} 
 }
\author{Qiong  Jia} 
\email{boajia@tencent.com}
\affiliation{  
	\institution{Tencent Youtu Lab} 
 	\city{Shanghai} 
	\country{ China} 
 }
\author{Quoc-Cuong Pham} 
\email{quoc-cuong.pham@cea.fr}
\affiliation{  
	\institution{Université Paris-Saclay, CEA, List} 
 	\city{Paris} 
	\country{ France} 
 }
\author{Ralph Ewerth} 
\email{ralph.ewerth@tib.eu}
\affiliation{  
	\institution{L3S Research Center  Leibniz University Hannover, \& TIB - Leibniz Information Center for Science and Technology} 
 	\city{Hannover} 
	\country{ Germany} 
 }
\author{Ran Song} 
\email{ransong@sdu.edu.cn}
\affiliation{  
	\institution{School of Control Science and Engineering, Shandong University} 
 	\city{Jinan} 
	\country{ China} 
 }
\author{Rengang Li} 
\email{lirg@inspur.com}
\affiliation{  
	\institution{Inspur Electronic Information Industry Co., Ltd. State Key Laboratory of High-end Server Storage Technology} 
 	\city{Jinan} 
	\country{ China} 
 }
\author{Rikke Gade} 
\email{rg@create.aau.dk}
\affiliation{  
	\institution{Aalborg University} 
 	\city{Aalborg} 
	\country{ Denmark} 
 }
\author{Ruben Debien} 
\email{ruben.debien@rebatch.be}
\affiliation{  
	\institution{ReBatch} 
 	\city{Kontich} 
	\country{ Belgium} 
 }
\author{Runze Zhang} 
\email{zhangrunze@inspur.com}
\affiliation{  
	\institution{Inspur Electronic Information Industry Co., Ltd. State Key Laboratory of High-end Server Storage Technology} 
 	\city{Jinan} 
	\country{ China} 
 }
\author{Sangrok Lee} 
\email{lsrock1@yonsei.ac.kr}
\affiliation{  
	\institution{Graduate school of information  yonsei university, \& MODULABS} 
 	\city{Seoul} 
	\country{ Korea} 
 }
\author{Sergio Escalera} 
\email{sergio.escalera.guerrero@gmail.com}
\affiliation{  
	\institution{Universitat de Barcelona, \& Computer Vision Center, \& Aalborg University} 
 	\city{Barcelona} 
	\country{ Spain} 
 }
\author{Shan Jiang} 
\email{jiang.shan@fujitsu.com}
\affiliation{  
	\institution{Fujitsu Research} 
 	\city{Kawasaki} 
	\country{ Japan} 
 }
\author{Shigeyuki Odashima} 
\email{sodashima@fujitsu.com}
\affiliation{  
	\institution{Fujitsu Research} 
 	\city{Kawasaki} 
	\country{ Japan} 
 }
\author{Shimin Chen} 
\email{chenshimin1@oppo.com}
\affiliation{  
	\institution{OPPO Research Institute} 
 	\city{Shenzhen } 
	\country{ China} 
 }
\author{Shoichi Masui} 
\email{masui.shoichi@fujitsu.com}
\affiliation{  
	\institution{Fujitsu Research} 
 	\city{Kawasaki} 
	\country{ Japan} 
 }
\author{Shouhong  Ding} 
\email{ericshding@tencent.com}
\affiliation{  
	\institution{Tencent Youtu Lab} 
 	\city{Shanghai} 
	\country{ China} 
 }
\author{Sin-wai Chan} 
\email{chansinwai@cihe.edu.hk}
\affiliation{  
	\institution{Caritas Institute of Higher Education} 
 	\city{Tseung Kwan O} 
	\country{ Hong Kong,  SAR} 
 }
\author{Siyu Chen} 
\email{chensiyu25@meituan.com}
\affiliation{  
	\institution{Meituan Inc.} 
 	\city{Beijing} 
	\country{ China} 
 }
\author{Tallal El-Shabrawy} 
\email{tallal.el-shabrawy@guc.edu.eg}
\affiliation{  
	\institution{German University in Cairo} 
 	\city{New Cairo City} 
	\country{ Egypt} 
 }
\author{Tao He} 
\email{kevin.92.he@gmail.com}
\affiliation{  
	\institution{Tsinghua University} 
 	\city{Beijing} 
	\country{ China} 
 }
\author{Thomas B. Moeslund} 
\email{tbm@create.aau.dk}
\affiliation{  
	\institution{Aalborg University} 
 	\city{Aalborg} 
	\country{ Denmark} 
 }
\author{Wan-Chi Siu} 
\email{enwcsiu@polyu.edu.hk}
\affiliation{  
	\institution{Caritas Institute of Higher Education, \& Hong Kong Polytechnic University} 
 	\city{Tseung Kwan O} 
	\country{ Hong Kong,  SAR} 
 }
\author{Wei Zhang} 
\email{davidzhang@sdu.edu.cn}
\affiliation{  
	\institution{School of Control Science and Engineering, Shandong University} 
 	\city{Jinan} 
	\country{ China} 
 }
\author{Wei Li} 
\email{liwei19@oppo.com}
\affiliation{  
	\institution{OPPO Research Institute} 
 	\city{Shenzhen } 
	\country{ China} 
 }
\author{Xiangwei Wang} 
\email{wangxiangwei@baidu.com}
\affiliation{  
	\institution{Department of Augmented Reality Technology (ART),  Baidu Inc.} 
 	\city{Beijing} 
	\country{ China} 
 }
\author{Xiao Tan} 
\email{tanxiao01@baidu.com}
\affiliation{  
	\institution{Department of Computer Vision Technology (VIS),  Baidu Inc.} 
 	\city{Beijing} 
	\country{ China} 
 }
\author{Xiaochuan Li} 
\email{lixiaochuan@inspur.com}
\affiliation{  
	\institution{Inspur Electronic Information Industry Co., Ltd. State Key Laboratory of High-end Server Storage Technology} 
 	\city{Jinan} 
	\country{ China} 
 }
\author{Xiaolin Wei} 
\email{weixiaolin02@meituan.com}
\affiliation{  
	\institution{Meituan Inc.} 
 	\city{Beijing} 
	\country{ China} 
 }
\author{Xiaoqing Ye} 
\email{yexiaoqing@baidu.com}
\affiliation{  
	\institution{Department of Computer Vision Technology (VIS), Baidu Inc.} 
 	\city{Beijing} 
	\country{ China} 
 }
\author{Xing Liu} 
\email{liuxing12@baidu.com}
\affiliation{  
	\institution{Department of Augmented Reality Technology (ART),  Baidu Inc.} 
 	\city{Beijing} 
	\country{ China} 
 }
\author{Xinying Wang} 
\email{xinying@mgtv.com}
\affiliation{  
	\institution{MGTV} 
 	\city{Changsha} 
	\country{ China} 
 }
\author{Yandong Guo} 
\email{guoyandong@oppo.com}
\affiliation{  
	\institution{OPPO Research Institute} 
 	\city{Shenzhen } 
	\country{ China} 
 }
\author{Yaqian Zhao} 
\email{zhaoyaqian@inspur.com}
\affiliation{  
	\institution{Inspur Electronic Information Industry Co., Ltd. State Key Laboratory of High-end Server Storage Technology} 
 	\city{Jinan} 
	\country{ China} 
 }
\author{Yi Yu} 
\email{yuyi@mgtv.com}
\affiliation{  
	\institution{MGTV} 
 	\city{Changsha} 
	\country{ China} 
 }
\author{Yingying Li} 
\email{liyingying05@baidu.com}
\affiliation{  
	\institution{Department of Computer Vision Technology (VIS), Baidu Inc.} 
 	\city{Beijing} 
	\country{ China} 
 }
\author{Yue He} 
\email{heyue04@baidu.com}
\affiliation{  
	\institution{Department of Computer Vision Technology (VIS), Baidu Inc.} 
 	\city{Beijing} 
	\country{ China} 
 }
\author{Yujie Zhong} 
\email{zhongyujie@meituan.com}
\affiliation{  
	\institution{Meituan Inc.} 
 	\city{Beijing} 
	\country{ China} 
 }
\author{Zhenhua Guo} 
\email{guozhenhua@inspur.com}
\affiliation{  
	\institution{Inspur Electronic Information Industry Co., Ltd. State Key Laboratory of High-end Server Storage Technology} 
 	\city{Jinan} 
	\country{ China} 
 }
\author{Zhiheng Li} 
\email{zhihengli@mail.sdu.edu.cn}
\affiliation{  
	\institution{School of Control Science and Engineering, Shandong University} 
 	\city{Jinan} 
	\country{ China} 
 }

\renewcommand{\shortauthors}{Silvio Giancola et. al.}

\begin{abstract}
The SoccerNet 2022 challenges were the second annual video understanding challenges organized by the SoccerNet team. In 2022, the challenges were composed of $6$ vision-based tasks: (1) action spotting, focusing on retrieving action timestamps in long untrimmed videos, (2) replay grounding, focusing on retrieving the live moment of an action shown in a replay, (3) pitch localization, focusing on detecting line and goal part elements, (4) camera calibration, dedicated to retrieving the intrinsic and extrinsic camera parameters, (5) player re-identification, focusing on retrieving the same players across multiple views, and (6) multiple object tracking, focusing on tracking players and the ball through unedited video streams. Compared to last year's challenges, tasks (1-2) had their evaluation metrics redefined to consider tighter temporal accuracies, and tasks (3-6) were novel, including their underlying data and annotations. More information on the tasks, challenges and leaderboards are available on \url{https://www.soccer-net.org}. Baselines and development kits are available on \url{https://github.com/SoccerNet}.
\renewcommand{\thefootnote}{\fnsymbol{footnote}}
\footnotetext[1]{Equal contributions (\url{silvio.giancola@kaust.edu.sa}, \url{anthony.cioppa@uliege.be}).}
\end{abstract}




\maketitle

\section{Introduction}\label{sec:introduction}

The topic of video understanding drew a lot of attention in computer vision research. 
In order to push research towards better video analysis tools in sports, the SoccerNet dataset introduces six tasks related to video understanding, which are supported by open challenges for the community.
This paper presents the final results of the SoccerNet 2022 challenges and gives voice to the participants who briefly present their solution.

\begin{figure}[htb]
  \includegraphics[width=\linewidth]{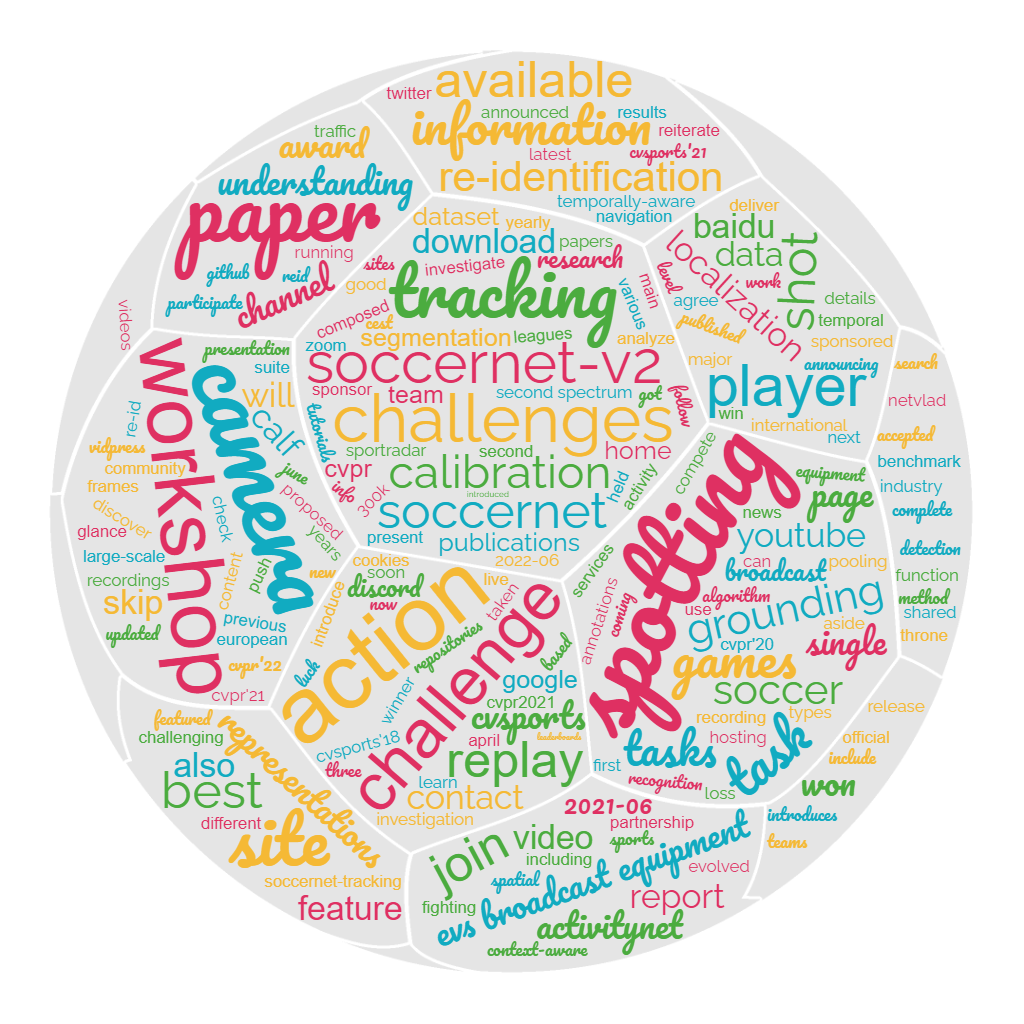}
  \caption{Word cloud generated from the word occurrence on the SoccerNet website. 
  One can spot the different tasks, baselines, sponsors, communication channels, and venues that are parts of the SoccerNet 2022 challenges.}
  \Description{Word cloud generated from the SoccerNet website.} 
  \label{fig:graphicalabstract}
\end{figure}

\subsection{SoccerNet dataset}

Giancola et al.~\cite{Giancola2018SoccerNet} introduced SoccerNet in 2018. The objective was to share a large-scale dataset for reproducible research in soccer video understanding and to define a new task of action spotting for the temporal localization of sports activities defined with single timestamps. Originally, the dataset contained $500$ videos of complete broadcast soccer games, totalling almost $800$ hours of videos from the six major European championships (Seria A, La Liga, Premier League, Ligue 1, Bundesliga, and Champion's League) from 2014 to 2017. The first annotations cover temporal timestamps of three main actions in soccer: goal, cards, and substitutions. The annotations were scrapped from websites with a one minute resolution and later manually refined to a one second precision.

Later, Deliège et al.~\cite{Deliege2021SoccerNetv2} introduced SoccerNet-v2, which significantly increased the number of annotations for the action spotting task by completing the set with common actions in soccer such as penalties, clearances, ball out of play, etc., for a total of $110{,}458$ actions split into $17$ classes. In addition to these extended annotations, SoccerNet-v2 integrates annotations for all camera changes among $13$ camera classes and three transition classes: abrupt, smooth, or logo. Finally, each camera shot is annotated by specifying if it contains a replay of an action, and if so, links it to its corresponding action timestamp during the live feed. Three tasks were proposed with this dataset: action spotting, camera shot segmentation and boundary detection, and replay grounding. Challenges were organized in 2021 for action spotting and replay grounding, the two tasks focusing on retrieving specific moments in videos, using $50$ extra games with segregated annotation as challenge set. 
    
This year, Cioppa et al.~\cite{Cioppa2022Scaling} introduced SoccerNet-v3, scaling up previous efforts by introducing spatial annotations and tasks. SoccerNet-v3 leveraged the redundancy of the actions from SoccerNet-v2 that were shown in both live and replay moments of the broadcast, and introduced spatial annotations for players, ball, field lines and goal parts from multiple views of the same scene. The frames corresponding to live and replayed actions were manually synchronized to the same salient moment of the action, totaling $33{,}986$ frames. Three novel tasks are defined based on those frames with spatial annotations. First, a pitch localization task that aims to recover semantic pitch elements such as the field lines and goal parts. Second, a camera calibration task aiming at estimating the intrinsic and extrinsic camera parameters, and finally, a player re-identification task focusing on retrieving the same player across multiple camera views. 
    
As the latest release, SoccerNet-Tracking~\cite{Cioppa2022SoccerNetTracking} introduces spatio-temporal annotations on a new set of $12$ complete games captured from a single main camera. The dataset includes $200$ $30$-seconds long clips extracted around key actions and a complete 45-minute half-time for long-term tracking with all objects annotated with bounding boxes, tracklet IDs, jersey numbers, and team tags. SoccerNet-Tracking is one of the largest multi-object tracking dataset and the largest one related to soccer, accounting for more than $3.6$~M bounding boxes and more than $5{,}000$ unique tracklets.

\subsection{SoccerNet challenges}

In the 2022 edition of the SoccerNet challenges, we proposed $6$ vision-based tasks: (1) action spotting, focusing on retrieving action timestamps in long untrimmed videos, (2) replay grounding, focusing on retrieving the live moment of an action shown in a replay, (3) pitch localization, focusing on detecting line and goal part elements, (4) camera calibration, focusing on retrieving the intrinsic and extrinsic camera parameters, (5) player re-identification, focusing on retrieving the same players across multiple views, and (6) multiple object tracking, focusing on tracking players and the ball through unedited video streams. 
%
%
The data for each challenge is split in $4$ sets: a training set and a validation set for training models, a public test set for benchmarking in scientific publications, and a private challenge set for ranking participants, whose annotations are kept segregated to avoid any cheating. 

To help participants get started with these challenges, we provide sample code on different SoccerNet GitHub repositories (\url{https://github.com/SoccerNet}) to download the data, run task-specific baselines, and evaluate their performance.

To facilitate interactions between participants, we created a Discord server, gathering more than $300$ researchers during the 2022 challenges. Furthermore, we organized several live tutorials with Q\&As and published explanatory videos on YouTube (\url{https://www.youtube.com/c/acadresearch}) to further attract the interest of the community.
In total, $67$ teams competed on the $6$ proposed tasks and submitted $637$ results files. 
We offered prizes for the winner of each task sponsored by 
Sportradar ($2{,}000~$\$ for tasks 1-2\&5),
EVS Broadcast Equipment ($1{,}000~$\$ for tasks 3-4), and 
Baidu Research ($1{,}000~$\$ for task 6).

In the following, we present a detailed analysis of each task, including its description and metric, the final leaderboard, a presentation of the best performing method by the team itself, and an analysis of the results. Each other participant team was entitled to present its method in the Appendix.
\section{Action Spotting}
\label{sec:spotting}

\subsection{Task description}

Action spotting can be considered one of the highest level of understanding for a soccer broadcast. It consists of localizing temporally when specific actions of interest occur (\eg penalty, kick-off, goal, etc.). Unlike other temporal localization tasks in video understanding (\eg temporal activity localization), the actions to spot are defined with single timestamps, based on soccer rules. For example, a goal is defined as the exact timestamp the ball crosses the goal line and a corner as the precise moment the player kicks the ball from the corner of the field.

Spotting soccer actions can be the building block of several applications in soccer video understanding, such as automatic video summarization and salient moment retrieval in live broadcasts. Furthermore, in its lowest level of granularity, it can support the generation of extended statistics for players and teams.

In this year's challenge, we leveraged the videos and annotations from SoccerNet-v2~\cite{Deliege2021SoccerNetv2}. The data consists of $500$ games, each of them split into two half-time videos of $45$ min plus eventual extra time. 
The annotations amount to $110{,}458$ actions from $17$ classes, anchored with a single timestamp.
In addition to these annotated data, we reserved extra $50$ games for the scope of this challenge, with segregated annotations to impede any participant team to train or overfit on this set.


\subsection{Metrics}

We use the Average-mAP~\cite{Giancola2018SoccerNet} metric for action spotting.
A predicted action spot is considered as a true positive if it falls within
a given tolerance $\delta$ of a ground-truth timestamp from the same class. 
The Average Precision (AP) based on PR curves is computed then averaged over the classes (mAP), after what the Average-mAP is the AUC of the mAP computed at different tolerances $\delta$. 
We define the \textit{loose Average-mAP} using the original tolerances $\delta$ ranging from $5$ to $60$ seconds~\cite{Giancola2018SoccerNet}.
We introduce a novel \textit{tight Average-mAP} with stricter tolerances $\delta$ ranging from $1$ to $5$ seconds, to evaluate for a more precise spotting.

Moreover, we differentiate between actions that are visible in the broadcast video, versus the actions that are not directly shown. For instance, several throw-ins and indirect free-kicks are not shown in the broadcast but can still be inferred from the dynamic of a game, after a ball went out of play or after a foul occurred. Spotting unshown actions requires a more abstract level of understanding involving the learning of causality and game logic.


\subsection{Leaderboard}

This year, $19$ teams participated to the action spotting challenge for a total of $167$ submissions, with an improvement from $49.56$ to $67.81$ tight Average-mAP. The leaderboard reporting the top-3 perfomances may be found in Table~\ref{tab:ShortLeaderboardSpotting}.


\begin{table}[t]
    \caption{Top-3 action spotting leaderboard, complete leaderboard available in Table~\ref{tab:LeaderboardSpotting} in the appendix. Main metric for the leaderboard and best performances in bold. Team names with a superscript have provided a summary that may be found in Appendix~\ref{app:spotting} or in Section~\ref{sub:spottingwinner} for the winning team.
    }
    \label{tab:ShortLeaderboardSpotting}
    \centering
\resizebox{\linewidth}{!}{
    \begin{tabular}{l||c|c|c||c|c|c}
\multirow{ 2}{*}{Participants} & \multicolumn{3}{c||}{\textbf{tight Average-mAP}} & \multicolumn{3}{c}{loose Average-mAP} \\ \cline{2-7}
& \textbf{main} & vis. & inv. & main & vis. & inv. \\ \midrule
\bf{Yahoo Research$^{S1}$} & \bf{67.81} & 72.84  & 60.17 & 78.05  & 80.61 & 78.05 \\
PTS & 66.73 & 74.84  & 53.21 & 73.62  & 79.16 & 67.42 \\
AS\&RG$^{S3}$ & 64.88 & 70.31  & 53.03 & 72.83  & 76.08 & 72.35 \\
Baseline* & 49.56* & 54.42  & 45.42 & 74.84  & 78.58 & 71.52 \\

    \end{tabular}
}
\end{table}

\subsection{Winner}
\label{sub:spottingwinner}

The winners for this task are João Soares \etal from the Yahoo Research, USA. A summary of their method is given hereafter.

\mysection{S1 - Dense Detection Anchors}\\
\textit{Jo{\~a}o V. B. Soares and Avijit Shah}\\
\textit{jvbsoares@yahooinc.com, avijit.shah@yahooinc.com}

Soares et al.~\cite{soares2022temporally} proposed an anchor-based approach, defining an anchor as a pair formed by a time instant and an action class, with time instants sampled densely. For each anchor, both a detection confidence and a fine-grained temporal displacement were inferred, with the displacement indicating exactly when an action was predicted to happen. The approach resulted in a substantial improvement to temporal precision, reaching 60.7 tight average-mAP. Specifically for the challenge, changes were introduced that led to the final 67.8 tight average-mAP on the challenge set, as detailed in a follow-up report~\cite{soares2022action}. While their method uses pre-computed features, for the challenge, two different feature types (Baidu and ResNet) were combined using a standard late fusion approach, after resampling them to the desired temporal frequency of two feature vectors per second. In addition, they applied a soft version of non-maximum suppression for post-processing, while optimizing the corresponding suppression window size.

\subsection{Results}

This year's challenge participants focused on improving video encoders and spotting heads. 
The video encoders evolved from CNN to transformers, learning spatial and/or temporal self-attention mechanisms. Some methods investigated multi-modality reasoning with additional audio encoders. The spotting heads were mostly adapted from temporal activity localization methods, with dense detection anchors and hierarchical action grouping.

It is worth noting that the leading method in tight Average-mAP (Yahoo Research) also performs best in the loose metric. However, on the subset of visible actions, the 2nd best method (PTS) outperforms the leader. We believe PTS primarly relies on visual cues while Yahoo Research's method has a deeper understanding of soccer rules, with best results on actions unshown in the broadcasts.




\section{Replay Grounding}
\label{sec:grounding}

\subsection{Task description}

Replays of salient moments are regularly shown in broadcast soccer games to emphasis the importance of an action, visualized under a more informative angle. Being able to link replays with their corresponding actions is thus a great tool for ranking actions by their impact on the game, which may be used to generate highlights of the game.

Given a replay clip, the goal of the replay grounding task is to spot the same action during the live game. The action timestamp correspond to the ones of the action spotting task, and thus follow the same annotation format.
Thus, the dataset consists of the same $500$ broadcast games from the action spotting task from which all replays have been retrieved. An extra $50$ games with segregated annotations compose the challenge set.

\subsection{Metrics}

The replay grounding task may be viewed as retrieving a single timestamp in a long untrimmed video. Hence, the same metrics as the ones used for the action spotting challenge may be used for this task. However, unlike action spotting, replay grounding does not consider the action class in its evaluation. Hence both the tight and loose average mean-Average Precision metrics are adapted by removing the averaging over the classes. These new metrics are called the tight and loose Average-AP.

For the tight Average-AP, we consider intervals of $1$ to $5$ seconds with a step of $1$ seconds, and for the loose Average-AP, we consider intervals of from $5$ to $60$ seconds with a step of $5$ seconds, following the action spotting metrics.

\subsection{Leaderboard}

This year, a single team submitted results on the replay grounding challenge set. Their performance may be found in Table~\ref{tab:LeaderboardGrounding}, alongside the baseline performance.

\begin{table}[t]
    \caption{Replay grounding leaderboard. Main metric for the leaderboard and best performances in bold. The winning team summary may be found in Section~\ref{sub:groundingwinner}. The baseline description may be found in \url{https://github.com/SoccerNet/sn-grounding}.}
    \label{tab:LeaderboardGrounding}
    
    \centering
    \begin{tabular}{l||c|c||c|c}
\multirow{ 2}{*}{Participants} & \multicolumn{2}{c||}{\textbf{tight Average-AP}} & \multicolumn{2}{c}{loose Average-AP} \\ \cline{2-5}
& \textbf{Challenge} & Test & Challenge & Test \\ \midrule
\bf{AS\&RG$^{G1}$} & \bf{45.33} & 52.31 & 	61.07  & 68.57 \\
Baseline* & 19.12* & 25.55 & 71.90 & 76.00 
    \end{tabular}
\end{table}

\subsection{Winner}
\label{sub:groundingwinner}

The winners for this task are Shimin Chen \etal from the OPPO Research Institute, China. A summary of their method is given hereafter.

\mysection{G1 - Video Action Location}\\
\textit{Shimin Chen, Wei Li, Jiaming Chu, Chen Chen, Chen Zhang, and Yandong Guo}\\
\textit{chenshimin1@oppo.com, liwei19@oppo.com, \\chujiaming886@bupt.edu.cn, chenchen@oppo.com, \\zhangchen4@oppo.com, guoyandong@oppo.com}

In order to make full use of video information, we transform the replay grounding problem into a video action location problem. We select 120 seconds clip before replay timestamps as input clip, and we set the timestamp label as the starting second of the segment labels with 3 seconds length. In this way, the predicted live stream timestamp corresponding to replay moment is equivalent to the start position of our detected result. As for temporal action detection, we first train VideoSwinTransformer~\cite{Liu2021VideoST} to extract video features. Then, we apply a unified network Faster-TAD~\cite{chen2022faster} proposed by us to get segments. To get more samples for training, we randomly synthesize positive samples. Finally, by observing the data distribution of the training data, we refine results to get the final submission. Our method reached a tight mAP of 52.31\% in test of SoccerNet Challenge 2022, bringing a gain of 26.76\% mAP relative to last year’s top result.

\subsection{Results}

The baseline performance correspond to last year's winner~\cite{Zhou2021FeatureCM}. As shown in Table~\ref{tab:LeaderboardGrounding}, this year's winning method significantly improved the spotting performance for tight intervals in both the challenge and test sets. These results show that the temporal activity module has a much better localization capability compared to the baseline. However, the loose average-AP significantly drops compared to the baseline. This may be due to the fact that the winner's method also makes several other guesses with high confidence, while the baseline usually focuses on a single instant, even though it is not perfectly localized.

\section{Field Localization}
\label{sec:FieldLocalization}

\subsection{Task description}
In the context of live sports events, camera calibration has many applications. One of them is to insert graphics in augmented reality for storytelling or to enforce the rules of the game (\eg drawing the offside line). The automatic calibration of a camera can be done leveraging correspondences between a known 3D representation of the scene, named a calibration pattern, and its image. In soccer, the field has a specific shape and appearance, which makes it a convenient calibration pattern. Therefore, in order to achieve camera calibration, we propose a first task consisting in the localization of the soccer field elements in the image.

Given an image, the goal of the field localization task is to detect each class of soccer field element present in the image, and also to predict the 2D points in the image representing the extremities of every soccer field element detected. The soccer field elements are the set of soccer field line or circle markings, and the three posts constituting each goal. Note that the extremity of an element is defined as either its true end, or the intersection of the object with the border of the image. 

The dataset has been annotated with polylines, a sequential list of 2D points that fits any soccer field element of rectilinear or circular nature. In this task, the objective is to retrieve the first and the last element, \ie the extremities, of each annotated polyline.

\subsection{Metrics}

\global\long\def\AE{AF}%

As there might be some uncertainty on the true exact location of an extremity, we threshold the Euclidean distance between a predicted extremity and its corresponding annotation in order to assess its validity. This thresholding strategy allows us to frame the problem as a detection task that can be evaluated by an accuracy metric dependent on the threshold value ($t$). We evaluate the predictions at different threshold levels. Concretely, we define that a point $x$ belonging to the predictions of class $C$ is a true positive ($TP$) if : $ x \in TP : \min_i \left\Vert x, \hat{x}_i \right \Vert_2 < t$ with ${\hat{x}_i}$ being the set of extremities annotated for the class $C$ in the image. The predicted extremities that do not meet that condition are counted as false positives ($FP$), along with predicted extremities that do not have a matching class in the annotations. Lastly, the false negatives ($FN$) are the extremities present in the annotations unmatched with any prediction. We define the Accuracy of the Field localization task within a tolerance of $t$ pixels $\AE@t$ as: $\AE@t = \frac{TP}{TP+FP+FN}$. The final evaluation is a weighted sum defined as $ 0.5\, \AE@5 + 0.35 \,\AE@10 + 0.15\, \AE@20.$

\subsection{Leaderboard}

\begin{table}[t]
\caption{Top-3 field localization leaderboard, complete leaderboard available in Table~\ref{tab:LeaderboardFieldLocalization} in the appendix. Main metric for the leaderboard and best performance in bold. Team names with a superscript provided a summary that can be found in Appendix \ref{app:pitch}, or in Section \ref{sub:fieldlocwinner} for the winner\label{tab:ShortLeaderboardFieldLocalization}.}

\begin{tabular}{l||c|c|c||c}
Participants & \AE@5 & \AE@10 & \AE@20 & \textbf{Final score}\tabularnewline
\midrule 
ONEDAY$^{P1}$ & 84.40 & 90.24 & 92.17 & \textbf{87.61}\tabularnewline
\hline 
imgo$^{P2}$ & 74.19 & 84.59 & 87.62 & 79.84\tabularnewline
\hline 
2Ai-IPCA$^{P3}$ & 71.01 & 76.18 & 77.60 & 73.81\tabularnewline
\hline 
Baseline* & 13.32 & 38.28 & 53.87 & 28.14*\tabularnewline
\end{tabular}

\end{table}

For this first edition of the field localization challenge, $12$ teams competed on the challenge set, for a total of $163$ submissions. The top-3
performances are reported in Table~\ref{tab:ShortLeaderboardFieldLocalization}.

\subsection{Winner}
\label{sub:fieldlocwinner}

The winners for this task are Yue He \etal from Baidu Inc, China. A summary of their method is given hereafter.

\mysection{P1 - Pitch Localization Detector (PLD)}\\
\textit{Yue He, Xiangwei Wang, Xing Liu, Xiaoqing Ye, Yingying Li, Chen Zhao, and Xiao Tan}\\
\textit{heyue04@baidu.com, wangxiangwei@baidu.com, \\liuxing12@baidu.com, yexiaoqing@baidu.com, \\liyingying05@baidu.com, zhaochen03@baidu.com, \\tanxiao01@baidu.com}

The task evaluation is dependent on the distance for the various class lines extremities. Besides, we observe that each line is unique, that is, there is at most one instance of a category of objects for a given image from the soccer pitch. Therefore, we treat it as an instance segmentation task at first that can correctly handle occlusions where an object is spilled into two separate regions. In this way, we build the framework of Pitch Localization Detector (PLD) with a Mask2Former \cite{cheng2022masked}, a state-of-the-art universal image segmentation model to identify the lines category, and a PP-YOLOv2 \cite{huang2021pp} detection model for optimizing extremities locations followed with a series of optimization strategy steps which include refinement with point results, dealing with left-right ambiguities, merging intersection points, geometry-based check, and merging output results. Therefore, our PLD method predicts the extremities of the soccer pitch elements present in each image.

\subsection{Results}
As can be seen in Table \ref{tab:ShortLeaderboardFieldLocalization}, the winner team obtains a significant performance gain compared to other teams. It can be explained by their combination of two modalities, \ie soccer field element instance segmentation and extremities detection, whereas other participants relied on semantic segmentation only. Another differentiating factor between the winning team and other participants is the use of recent neural networks architectures, such as a transformer for the segmentation of soccer field elements. 
\section{Camera Calibration}
\label{sec:Camera Calibration}

\subsection{Task description}
As previously mentioned in Section~\ref{sec:FieldLocalization}, the automatic calibration of broadcast cameras is a game-changer to bring augmented reality graphics into live production. The goal of the task is to retrieve intrinsic and extrinsic camera parameters based on a single frame. The pinhole camera model is imposed, with some flexibility regarding the distortion parameters of the lens. Indeed, participants can choose to provide tangential, radial and thin prism distortion. 

Following the previous task, we provide a 3D model of the soccer field to allow the mapping of the extremities located in the previous task to the 3D points of the field. This 3D model is further used in the evaluation.

For this task, the annotations are the same as in the previous section, but this time we keep all the annotated points of the polylines whilst before, we selected only each polyline's extremities. We emphasize the absence of any ground-truth concerning the extrinsic and intrinsic camera parameters. The evaluation is only based on metrics measuring the reprojection error in the image.

\subsection{Metrics}

\global\long\def\AS{AC}%
\global\long\def\CR{CR}%

In order to assess the quality of a submission, we provide several metrics. First, we must take into account the fact that there are some calibration methods that will fail to provide results on certain images, which is why we introduce a ``Completeness Ratio'' (\CR) that is the ratio of the dataset images for which the method provides camera parameters. 
Then the other metrics are based on the accuracy of the projection of each soccer field element in the image. Using our provided soccer field model, we sample 3D points regularly along each soccer field element, then project each point in the image using the predicted camera parameters for a specific frame. In this way we obtain a set of 2D polylines that we can compare to the annotated polylines.
Given a point in the 3D world $\textbf{X}$ that has been sampled along a soccer field element of our 3D soccer field model, we use the predicted camera parameters to derive its projection in the image $\textbf{x}$. The projection first transforms the point $\textbf{X}$ to the camera reference system using the predicted rotation matrix $\text{R}$ and translation vector $\textbf{t}$ : $\left(X_c, Y_c, Z_c\right)^T = \left[\begin{array}{cc}
\text{R} & \textbf{t}\end{array}\right]\left(X,Y,Z\right)^T$. Then the point is projected in the normalized image plane : $\left(x',y'\right) = \left(\frac{X_c}{Z_c}, \frac{Y_c}{Z_c}\right) $, where distortion can be applied using the set of predicted distortion coefficient $r$ : $\left(x_d, y_d\right) = \psi_r(x', y')$ where $\psi_r$ is the function applying radial, tangential and thin prism distortion. Finally, we obtain the final pixel coordinates of $\textbf{x}$ using the predicted focal lengths $ f_x$ and $f_y$ as well as the principal point $\left(c_x, c_y\right)$  : $\left(x, y\right) = \left(f_x x_d + c_x, f_y y_d + c_y \right).$ 

The 2D point $\textbf{x}$ will be part of the 2D polyline associated with the class of the soccer field element. Our idea is again to frame this evaluation as a detection of soccer field elements in the image. 
We define that a polyline corresponding to a soccer field element $l$ is correctly detected if the Euclidean distance between every point belonging to the annotated polyline $\hat{l}$ and the projected polyline $l$ is less than $t$ pixels: $ \forall \hat{x} \in \hat{l} : \left \Vert \hat{x}, l \right\Vert _2 < t$.  We count each predicted soccer field element that meets this condition as true positives ($TP$), whilst a predicted soccer field element that is located at more than $t$ pixels from one of the annotated points for this primitive is counted as a false positive ($FP$), along with projected polylines that do not appear in the annotations. The false negatives ($FN$) are the polylines annotated that do not have a corresponding prediction. Finally, we define the Accuracy for the Camera calibration task within a tolerance of $t$ pixels as: $\AS@t = \frac{TP}{TP+FN+FP}$. We combine, in a weighted average, several levels of $\AS@t$ and we apply a trade-off between the completeness rate and this weighted average in order to produce our final evaluation metric. The idea of the trade-off is to encourage participants to focus on improving accuracy rather than robustness as the completeness ratio is increasing. This is ensured by the use of a factor containing a negative exponential of the completeness ratio: an improvement in a small completeness ratio value has a higher positive impact on the metric rather than the same improvement with already satisfying completeness rate.  
This yields the following final score $s$ defined as  
$s = (1- e^{-4 \, \text{\CR}})( 0.5\, \AS@5 + 0.35 \,\AS@10 + 0.15\, \AS@20).$

\subsection{Leaderboard}

\begin{table}[t]
\caption{Top-3 camera calibration leaderboard, complete leaderboard available in Table~\ref{tab:LeaderboardCalibration} in the appendix. Main metric for the leaderboard and
best performance in bold. Team names with a superscript provided a
summary that can be found in Appendix \ref{app:camera},
or in Section \ref{sub:camerawinner} for
the winner\label{tab:ShortLeaderboardCalibration}.}

\begin{tabular}{l||c|c|c|c||c}

Participants & \AS@5 & \AS@10 & \AS@20 & \CR & \textbf{Final $s$}\tabularnewline
 
\midrule 
achengmao$^{C1}$ & 82.38 & 94.80 & 96.33 & 72.61 & \textbf{83.96}\tabularnewline
\hline 
L3S$^{C2}$ & 57.83 & 81.42 & 90.74 & 69.32 & 66.58\tabularnewline
\hline 
MikeAzatov$^{C3}$ & 62.25 & 84.32 & 90.56 & 56.41 & 66.45\tabularnewline
\hline 
Baseline* & 12.94 & 29.14 & 43.48 & 58.95 & 21.00*\tabularnewline
\end{tabular}

\end{table}

For this first edition of the camera calibration challenge, $6$ teams competed on the challenge set, for a total of $63$ submissions. The top-3
performances are reported in Table~\ref{tab:ShortLeaderboardCalibration}.

\subsection{Winner}
\label{sub:camerawinner}

The winners for this task are Xiangwei Wang \etal from Baidu Inc, China. A summary of their method is given hereafter.

\mysection{C1 - Achengmao}\\
\textit{Xiangwei Wang, Xing Liu, Yue He, Xiaoqing Ye, Yingying Li, Chen Zhao, and Xiao Tan}\\
\textit{wangxiangwei@baidu.com, liuxing12@baidu.com, \\heyue04@baidu.com, yexiaoqing@baidu.com, \\liyingying05@baidu.com, zhaochen03@baidu.com, \\tanxiao01@baidu.com}

We address the problem of camera calibration for soccer videos. Given a frame extracted from a video, we detect and segment the elements (\eg, lines, conics) of the pitch. We compute five types of landmark, which are line-line intersection, conic-line intersection, field center, vanishing point, and points at curves based on the detection and segmentation results. To ensure accurate landmarks, we: (1) resolve ambiguities caused by the symmetric nature of soccer field, (2) prevent each pair of lines from incorrectly splitting into two from a whole; (3) reject incorrect conic-line intersections.  We propose three solvers to estimate the homograph for calibration in parallel. They are all points solver, RANSAC solver w/ and w/o coordinate perturbation. We determine the winner solver with the minimum re-projection error and conduct additional optimizations on it to obtain the optimal result of our method. The proposed method have achieved the first place in SoccerNet 2022 calibration competition.

\subsection{Results}
Since the algorithm provided for the previous task is used to solve the camera calibration problem, there is a strong dependency between the results obtained on the previous task and those achievable for the current task. It is therefore not surprising that with such a lead in the detection of football field features, the best camera calibration method is that of the best team on the previous task. In a later edition of this challenge, we will consider further disentanglement between the two tasks, in order to evaluate solely the calibration method without implicitly also evaluating the underlying semantic feature detection.

\section{Player Re-Identification}
\label{sec:reidentification}

\subsection{Task description}

Person re-identification \cite{Ye2022DeepLF}, or simply ReID, is a person retrieval task which aims at matching an image of a person-of-interest, called the \textit{query}, with other person images within a large database, called the \textit{gallery}, captured from various camera viewpoints. 
ReID has important applications in smart cities, video-surveillance and sport analytics, where it is used to perform person retrieval or tracking.

The goal of the SoccerNet ReID task is to re-identify players and referees across multiple camera viewpoints for a given action at a specific time instant during a soccer game.
Our SoccerNet re-identification dataset is composed of $340{,}993$ players thumbnails extracted from image frames of broadcast videos from $400$ soccer games within $6$ major leagues.
%
%

Compared to traditional street surveillance type re-identification dataset, the SoccerNet-v3 ReID dataset is particularly challenging because soccer players from the same team have very similar appearance, which makes it hard to tell them apart. 
On the other hand, each identity has a few amount of samples, which makes the model harder to train. 
Finally, there is a big diversity within samples of the dataset in terms of image resolution.


\subsection{Metrics}

We use two standard retrieval evaluation metrics to compare different ReID models: the cumulative matching characteristics (CMC) \cite{Wang2007ShapeAA} at Rank-1 and the mean average precision \cite{Zheng2015ScalablePR} (mAP).
Participants to the SoccerNet ReID challenge have been ranked according to their mAP score on the challenge set.

\subsection{Leaderboard}
For this first edition of the player ReID challenge, $13$ teams competed on the challenge set, for a total of $123$ submissions. Their top-3 performances are reported in Table \ref{tab:ShortLeaderboardReID}.


\begin{table}[htb]
\caption{Top-3 leaderboard for the ReID task, complete leaderboard available in Table~\ref{tab:LeaderboardReID} in the appendix. Main metric for the leaderboard and best performance in bold. Team names with a superscript have provided a summary that can be found in the appendix, or in the next section for the winner.}
\label{tab:ShortLeaderboardReID}
\centering
\begin{tabular}{l||c|c}
Participants & \textbf{mAP} & R-1 \\ \midrule
\textbf{Inspur$^{T1}$} & 91.68 & 89.41 \\
MGSoccer$^{T2}$ & 91.48 & 89.21 \\
MTVACV$^{T3}$   & 90.11 & 87.04 \\
Baseline        & 59.11 & 48.41
\end{tabular}
\end{table}

\subsection{Winner}
\label{sub:reidwinner}

The winners for this task are Rengang Li \etal from Inspur, China. A summary of their method is given hereafter.

\mysection{R1 - Optimized Strategy for Player Re-identification}\\
\textit{Rengang Li, Yaqian Zhao, Hongwei Kan, Zhenhua Guo, Baoyu Fan, Runze Zhang,
Xiaochuan Li}\\
\textit{lirg@inspur.com, zhaoyaqian@inspur.com, \\kanhongwei@inspur.com, guozhenhua@inspur.com, \\fanbaoyu@inspur.com, zhangrunze@inspur.com, \\lixiaochuan@inspur.com}

We analyzed that the main challenges are the sample imbalance and unrobustness mainly caused by multi-input resolution. We removed the ids whose images are less 3 and employed the focal loss function to solve sample imbalance. We experimented different combination of ReID network module to choose the best representation ability and selected ResNeSt269, combination of Arc-Softmax and Cos-Softmax. We used Auto-Aug, Color Jittering and Random Erase and all of the
data augmentation uses the probability of 0.5. After we optimized the best hyper-parameters of single model, we paid more attention to the common person ReID tricks, such as multi-input resolution model fusion, add test phase dataset as well as unsupervised domain adaptation.

\subsection{Results}

Participants came up with various innovative ideas and have achieved outstanding performances despite the difficulty of the task.
%
We list here some of the keys ideas shared by participants. 
\textbf{(i)} Apply some pre-processing by removing identities with too few samples in the training set.
\textbf{(ii)} Design a handcrafted training batch sampling strategies based on additional SoccerNet ReID dataset labels, such as action id and game id.
\textbf{(iii)} Add standard data augmentation strategies: Horizontal Flip, Random Erasing \cite{Zhong2020RandomED}, Random Cropping, AutoAugment \cite{Cubuk2018AutoAugmentLA}, AugMix \cite{Hendrycks2020AugMixAS}, Color Jitter, ...
\textbf{(iv)} Use a strong baseline such as the TransReID-SSL \cite{TransReid_SSL} baseline with ViT \cite{Dosovitskiy2021AnII} backbone and unsupervised pre-training on LUPerson \cite{LUPerson} dataset.
\textbf{(v)} Use specific metric learning loss functions: the Focal Loss \cite{Lin2020FocalLF}, a custom Centroid loss, the InfoNCE loss \cite{Oord2018RepresentationLW}, the Arcface loss \cite{Deng2019ArcFaceAA}, ...
\textbf{(vi)} Inference time fine-tuning with unsupervised domain adaptation on the challenge set to further increase final performance.
\textbf{(vii)} Combine multiple models predictions at inference to compute final distance metric.


\section{Multiple Player Tracking}
\label{sec:tracking}

\subsection{Task description}

Tracking is a hot topic of research, which is far from being solved.
In sports, tracking algorithms enable many interesting applications. They can be used to generate player specific highlights and statistics, or be leveraged for holistic video understanding~\cite{Cioppa2021Camera}. 

As defined in the SoccerNet-Tracking dataset, the tracking task is split in two steps: (1) detecting the objects to track and (2) associating the bounding boxes over time to create the tracklets.
For this year's challenge, the participants had access to $150$ $30-$seconds clips recorded only from a single camera, with all ground-truth bounding boxes provided. The goal of the task is therefore to associate these bounding boxes over time to create the final tracklets.
The complete tracking task, including both detection and association, will be part of the next edition of the SoccerNet challenges.

Compared to most tracking datasets, SoccerNet-Tracking includes several challenges such as long-term re-identification, \ie if an object leaves the frame and comes back, it needs to be associated to the same tracklet. Since most players in the same team have very similar appearances, the re-identification is challenging.

\subsection{Metrics}

Following the recent work of Luiten \etal~\cite{Luiten2021Hota}, we use the HOTA metric to rank the participants. This metric may be decomposed into a detection accuracy (DetA) and an association accuracy (AssA). Compared to the previous common MOTA metric, it is much more balanced for the evaluation of detection and association capabilities.

\subsection{Leaderboard}

For this first edition of the challenge, $12$ teams competed on the challenge set, for a total of $103$ submissions. The performance of the top-3 teams may be found in Table~\ref{tab:ShortLeaderboardTracking}.

\begin{table}[t]
    \caption{Top-3 tracking leaderboard, complete leaderboard available in Table~\ref{tab:LeaderboardTracking} in the appendix. Main metric for the leaderboard and best performances in bold. Team names with a superscript have provided a summary that may be found in Appendix~\ref{app:tracking}, or in Section~\ref{sub:trackingwinner} for the winning team.}
    \label{tab:ShortLeaderboardTracking}
    \centering
    \begin{tabular}{l||c|c|c}
Participants & \textbf{HOTA} & DetA & AssA \\ \midrule
\bf{Kalisteo$^{T1}$} & \bf{93.64} & 99.56 & 88.06 \\
CBIOUT (CB-IoU)$^{T2}$ & 93.25 & 99.76 & 87.15 \\
tactica$^{T3}$ & 93.17 & 99.85 & 86.94 \\
Baseline* & 	70.89* & 82.97 & 60.68 \\

    \end{tabular}
\end{table}

\subsection{Winner}
\label{sub:trackingwinner}

The winners for this task are Adrien Maglo~\etal from Université Paris-Saclay, CEA, List, France. A summary of their method is given hereafter.

\mysection{T1 - TrackMerger}\\
\textit{Adrien Maglo, Astrid Orcesi, and Quoc-Cuong Pham
}\\
\textit{adrien.maglo@cea.fr, astrid.orcesi@cea.fr, quoc-cuong.pham@cea.fr}

The first step of TrackMerger generates player tracks by sequentially processing the video frames. The current frame detections are matched to existing tracks bounding boxes with a Hungarian assignment algorithm using two criteria, the Intersection-Over-Union between bounding boxes and the distance between their center. Only small bounding boxes can extend the ball track.
Generated tracks are of good quality as long as the player stay visible. To be able to recognize players who exit and later re-enter the camera field of view, the second step fine-tunes a re-identification network with a triplet loss formulation. Positive samples are extracted from the same track as the anchor while negative samples come from concomitant tracks.
The third step merges the tracks according to the distance between their re-identification vectors. It also prevents the duplication of a player's identity in the same frame and teleportation in successive frames.

\subsection{Results}



Similar to the ReID challenge, participants achieved outstanding performances on this task.
Most participants used the standard two phases approach to address long-term tracking:
\textbf{(i) Short tracklets}: Build short tracklets using an online tracking method relying mainly on spatio-temporal features, such as IoU/BIoU with Kalman filter.
\textbf{(ii) Long tracks}: Connect these short tracklets in an offline manner using appearance features, in order to solve heavy occlusions or players going out of the camera view. These appearance features are obtained using pre-trained re-identification models, that are fine-tuned on the training set or that are learned at inference in a self-supervised way on the short tracklets generated in the previous step.
Some participants used additional priors to further improve HOTA performance, such as physical constraints on ball size or players maximum speed.

\section{Conclusion}\label{sec:conclusion}
This paper summarizes the outcome of the SoccerNet 2022 challenges. 
In total, we present the results on six tasks: action spotting, replay grounding, pitch localization, camera calibration, player re-identification, and player tracking. 
These challenges provide a comprehensive overview of current state-of-the-art methods within each computer vision task. 
For each challenge, participants were able to significantly improve the performance of our proposed baselines, introducing new architectures, engineering tricks, and soccer-centric priors.
Yet, much more effort is still needed to solve the proposed tasks for practical applications.
In future editions, we expect to enrich the current sets of annotations and propose further tasks related to video understanding in soccer, introducing multiple modalities, higher level of granularity, and summarization tasks. 


\begin{acks}
This work was supported by the Service Public de Wallonie (SPW) Recherche under the DeepSport project and Grant $\text{N}^\text{o}$. 2010235 (ARIAC by \url{https://DigitalWallonia4.ai}), the FRIA, the FNRS, and KAUST Office of Sponsored Research through the Visual Computing Center funding.
\end{acks}


\bibliographystyle{ACM-Reference-Format}
\balance
\bibliography{abbreviation-short,bibliography}


\begin{thebibliography}{32}


\ifx \showCODEN    \undefined \def \showCODEN     #1{\unskip}     \fi
\ifx \showDOI      \undefined \def \showDOI       #1{#1}\fi
\ifx \showISBNx    \undefined \def \showISBNx     #1{\unskip}     \fi
\ifx \showISBNxiii \undefined \def \showISBNxiii  #1{\unskip}     \fi
\ifx \showISSN     \undefined \def \showISSN      #1{\unskip}     \fi
\ifx \showLCCN     \undefined \def \showLCCN      #1{\unskip}     \fi
\ifx \shownote     \undefined \def \shownote      #1{#1}          \fi
\ifx \showarticletitle \undefined \def \showarticletitle #1{#1}   \fi
\ifx \showURL      \undefined \def \showURL       {\relax}        \fi
\providecommand\bibfield[2]{#2}
\providecommand\bibinfo[2]{#2}
\providecommand\natexlab[1]{#1}
\providecommand\showeprint[2][]{arXiv:#2}

\bibitem[Boiarov and Tyantov(2019)]%
        {boiarov2019large}
\bibfield{author}{\bibinfo{person}{Andrei Boiarov} {and}
  \bibinfo{person}{Eduard Tyantov}.} \bibinfo{year}{2019}\natexlab{}.
\newblock \showarticletitle{Large Scale Landmark Recognition via Deep Metric
  Learning}. In \bibinfo{booktitle}{\emph{ACM Int. Conf. Inf. Knowl. Manag.}}
  \bibinfo{publisher}{ACM}, \bibinfo{address}{Beijing China},
  \bibinfo{pages}{169--178}.
\newblock
\urldef\tempurl%
\url{https://doi.org/10.1145/3357384.3357956}
\showDOI{\tempurl}


\bibitem[Chen et~al\mbox{.}(2022)]%
        {chen2022faster}
\bibfield{author}{\bibinfo{person}{Shimin Chen}, \bibinfo{person}{Chen Chen},
  \bibinfo{person}{Wei Li}, \bibinfo{person}{Xunqiang Tao}, {and}
  \bibinfo{person}{Yandong Guo}.} \bibinfo{year}{2022}\natexlab{}.
\newblock \showarticletitle{{Faster-TAD}: Towards Temporal Action Detection
  with Proposal Generation and Classification in a Unified Network}.
\newblock \bibinfo{journal}{\emph{arXiv}}  \bibinfo{volume}{abs/2204.02674}
  (\bibinfo{year}{2022}), \bibinfo{numpages}{16}~pages.
\newblock
\showeprint{2204.02674}


\bibitem[Cheng et~al\mbox{.}(2022)]%
        {cheng2022masked}
\bibfield{author}{\bibinfo{person}{Bowen Cheng}, \bibinfo{person}{Ishan Misra},
  \bibinfo{person}{Alexander~G Schwing}, \bibinfo{person}{Alexander Kirillov},
  {and} \bibinfo{person}{Rohit Girdhar}.} \bibinfo{year}{2022}\natexlab{}.
\newblock \showarticletitle{Masked-attention mask transformer for universal
  image segmentation}. In \bibinfo{booktitle}{\emph{IEEE Conf. Comput. Vis.
  Pattern Recog.}} \bibinfo{address}{New Orleans, LA, USA},
  \bibinfo{pages}{1290--1299}.
\newblock


\bibitem[Cioppa et~al\mbox{.}(2022a)]%
        {Cioppa2022Scaling}
\bibfield{author}{\bibinfo{person}{Anthony Cioppa}, \bibinfo{person}{Adrien
  Deli{\`e}ge}, \bibinfo{person}{Silvio Giancola}, \bibinfo{person}{Bernard
  Ghanem}, {and} \bibinfo{person}{Marc Van~Droogenbroeck}.}
  \bibinfo{year}{2022}\natexlab{a}.
\newblock \showarticletitle{Scaling up {SoccerNet} with multi-view spatial
  localization and re-identification}.
\newblock \bibinfo{journal}{\emph{Scientific Data}} \bibinfo{volume}{9},
  \bibinfo{number}{1} (\bibinfo{date}{June} \bibinfo{year}{2022}),
  \bibinfo{pages}{1--9}.
\newblock
\urldef\tempurl%
\url{https://doi.org/10.1038/s41597-022-01469-1}
\showDOI{\tempurl}


\bibitem[Cioppa et~al\mbox{.}(2021)]%
        {Cioppa2021Camera}
\bibfield{author}{\bibinfo{person}{Anthony Cioppa}, \bibinfo{person}{Adrien
  Deli{\`e}ge}, \bibinfo{person}{Silvio Giancola}, \bibinfo{person}{Floriane
  Magera}, \bibinfo{person}{Olivier Barnich}, \bibinfo{person}{Bernard Ghanem},
  {and} \bibinfo{person}{Marc Van~Droogenbroeck}.}
  \bibinfo{year}{2021}\natexlab{}.
\newblock \showarticletitle{Camera Calibration and Player Localization in
  {SoccerNet-v2} and Investigation of their Representations for Action
  Spotting}. In \bibinfo{booktitle}{\emph{IEEE Int. Conf. Comput. Vis. and
  Pattern Recogn. Work. (CVPRW), CVsports}}. \bibinfo{publisher}{Inst. Elect.
  and Electron. Engineers (IEEE)}, \bibinfo{address}{Nashville, TN, USA},
  \bibinfo{pages}{4537--4546}.
\newblock
\urldef\tempurl%
\url{https://doi.org/10.1109/CVPRW53098.2021.00511}
\showDOI{\tempurl}


\bibitem[Cioppa et~al\mbox{.}(2022b)]%
        {Cioppa2022SoccerNetTracking}
\bibfield{author}{\bibinfo{person}{Anthony Cioppa}, \bibinfo{person}{Silvio
  Giancola}, \bibinfo{person}{Adrien Deli{\`e}ge}, \bibinfo{person}{Le Kang},
  \bibinfo{person}{Xin Zhou}, \bibinfo{person}{Cheng Zhiyu},
  \bibinfo{person}{Bernard Ghanem}, {and} \bibinfo{person}{Marc~Van
  Droogenbroeck}.} \bibinfo{year}{2022}\natexlab{b}.
\newblock \showarticletitle{{SoccerNet-Tracking}: Multiple Object Tracking
  Dataset and Benchmark in Soccer Videos}. In \bibinfo{booktitle}{\emph{IEEE
  Int. Conf. Comput. Vis. and Pattern Recogn. Work. (CVPRW), CVsports}}.
  \bibinfo{publisher}{Inst. Elect. and Electron. Engineers (IEEE)},
  \bibinfo{address}{New Orleans, LA, USA}, \bibinfo{pages}{3491--3502}.
\newblock


\bibitem[Comandur(2022)]%
        {Comandur2022SportsRI}
\bibfield{author}{\bibinfo{person}{Bharath Comandur}.}
  \bibinfo{year}{2022}\natexlab{}.
\newblock \showarticletitle{Sports Re-ID: Improving Re-Identification Of
  Players In Broadcast Videos Of Team Sports}.
\newblock \bibinfo{journal}{\emph{arXiv}}  \bibinfo{volume}{abs/2206.02373}
  (\bibinfo{year}{2022}), \bibinfo{numpages}{11}~pages.
\newblock
\showeprint{2206.02373}


\bibitem[Cubuk et~al\mbox{.}(2018)]%
        {Cubuk2018AutoAugmentLA}
\bibfield{author}{\bibinfo{person}{Ekin~Dogus Cubuk}, \bibinfo{person}{Barret
  Zoph}, \bibinfo{person}{Dandelion Man{\'e}}, \bibinfo{person}{Vijay
  Vasudevan}, {and} \bibinfo{person}{Quoc~V. Le}.}
  \bibinfo{year}{2018}\natexlab{}.
\newblock \showarticletitle{AutoAugment: Learning Augmentation Policies from
  Data}.
\newblock \bibinfo{journal}{\emph{arXiv}}  \bibinfo{volume}{abs/1805.09501}
  (\bibinfo{year}{2018}), \bibinfo{numpages}{14}~pages.
\newblock
\showeprint{1805.09501}


\bibitem[Deli{\`e}ge et~al\mbox{.}(2021)]%
        {Deliege2021SoccerNetv2}
\bibfield{author}{\bibinfo{person}{Adrien Deli{\`e}ge},
  \bibinfo{person}{Anthony Cioppa}, \bibinfo{person}{Silvio Giancola},
  \bibinfo{person}{Meisam~J. Seikavandi}, \bibinfo{person}{Jacob~V. Dueholm},
  \bibinfo{person}{Kamal Nasrollahi}, \bibinfo{person}{Bernard Ghanem},
  \bibinfo{person}{Thomas~B. Moeslund}, {and} \bibinfo{person}{Marc
  Van~Droogenbroeck}.} \bibinfo{year}{2021}\natexlab{}.
\newblock \showarticletitle{{SoccerNet}-v2: A Dataset and Benchmarks for
  Holistic Understanding of Broadcast Soccer Videos}. In
  \bibinfo{booktitle}{\emph{IEEE Int. Conf. Comput. Vis. and Pattern Recogn.
  Work. (CVPRW), CVsports}}. \bibinfo{publisher}{Inst. Elect. and Electron.
  Engineers (IEEE)}, \bibinfo{address}{Nashville, TN, USA},
  \bibinfo{pages}{4508--4519}.
\newblock
\urldef\tempurl%
\url{https://doi.org/10.1109/CVPRW53098.2021.00508}
\showDOI{\tempurl}


\bibitem[Deng et~al\mbox{.}(2019)]%
        {Deng2019ArcFaceAA}
\bibfield{author}{\bibinfo{person}{Jiankang Deng}, \bibinfo{person}{Jia Guo},
  \bibinfo{person}{Niannan Xue}, {and} \bibinfo{person}{Stefanos Zafeiriou}.}
  \bibinfo{year}{2019}\natexlab{}.
\newblock \showarticletitle{{ArcFace}: Additive Angular Margin Loss for Deep
  Face Recognition}. In \bibinfo{booktitle}{\emph{IEEE/CVF Conf. Comput. Vis.
  and Pattern Recogn. (CVPR)}}. \bibinfo{publisher}{Inst. Elect. and Electron.
  Engineers (IEEE)}, \bibinfo{address}{Long Beach, CA, USA},
  \bibinfo{pages}{4685--4694}.
\newblock
\urldef\tempurl%
\url{https://doi.org/10.1109/cvpr.2019.00482}
\showDOI{\tempurl}


\bibitem[Dosovitskiy et~al\mbox{.}(2021)]%
        {Dosovitskiy2021AnII}
\bibfield{author}{\bibinfo{person}{Alexey Dosovitskiy}, \bibinfo{person}{Lucas
  Beyer}, \bibinfo{person}{Alexander Kolesnikov}, \bibinfo{person}{Dirk
  Weissenborn}, \bibinfo{person}{Xiaohua Zhai}, \bibinfo{person}{Thomas
  Unterthiner}, \bibinfo{person}{Mostafa Dehghani}, \bibinfo{person}{Matthias
  Minderer}, \bibinfo{person}{Georg Heigold}, \bibinfo{person}{Sylvain Gelly},
  \bibinfo{person}{Jakob Uszkoreit}, {and} \bibinfo{person}{Neil Houlsby}.}
  \bibinfo{year}{2021}\natexlab{}.
\newblock \showarticletitle{An Image is Worth 16x16 Words: Transformers for
  Image Recognition at Scale}.
\newblock \bibinfo{journal}{\emph{arXiv}}  \bibinfo{volume}{abs/2010.11929}
  (\bibinfo{year}{2021}), \bibinfo{numpages}{22}~pages.
\newblock
\showeprint{2010.11929}


\bibitem[Feichtenhofer et~al\mbox{.}(2019)]%
        {feichtenhofer2019slowfast}
\bibfield{author}{\bibinfo{person}{Christoph Feichtenhofer},
  \bibinfo{person}{Haoqi Fan}, \bibinfo{person}{Jitendra Malik}, {and}
  \bibinfo{person}{Kaiming He}.} \bibinfo{year}{2019}\natexlab{}.
\newblock \showarticletitle{{SlowFast} Networks for Video Recognition}. In
  \bibinfo{booktitle}{\emph{Int. Conf. Comput. Vis.}} \bibinfo{publisher}{Inst.
  Elect. and Electron. Engineers (IEEE)}, \bibinfo{address}{Seoul, South
  Korea}, \bibinfo{pages}{6201--6210}.
\newblock
\urldef\tempurl%
\url{https://doi.org/10.1109/iccv.2019.00630}
\showDOI{\tempurl}


\bibitem[Fu et~al\mbox{.}(2021)]%
        {LUPerson}
\bibfield{author}{\bibinfo{person}{Dengpan Fu}, \bibinfo{person}{Dongdong
  Chen}, \bibinfo{person}{Jianmin Bao}, \bibinfo{person}{Hao Yang},
  \bibinfo{person}{Lu Yuan}, \bibinfo{person}{Lei Zhang},
  \bibinfo{person}{Houqiang Li}, {and} \bibinfo{person}{Dong Chen}.}
  \bibinfo{year}{2021}\natexlab{}.
\newblock \showarticletitle{Unsupervised Pre-training for Person
  Re-identification}. In \bibinfo{booktitle}{\emph{IEEE/CVF Conf. Comput. Vis.
  and Pattern Recogn. (CVPR)}}. \bibinfo{publisher}{Inst. Elect. and Electron.
  Engineers (IEEE)}, \bibinfo{address}{Nashville, TN, USA},
  \bibinfo{pages}{14745--14754}.
\newblock
\urldef\tempurl%
\url{https://doi.org/10.1109/cvpr46437.2021.01451}
\showDOI{\tempurl}


\bibitem[Giancola et~al\mbox{.}(2018)]%
        {Giancola2018SoccerNet}
\bibfield{author}{\bibinfo{person}{Silvio Giancola},
  \bibinfo{person}{Mohieddine Amine}, \bibinfo{person}{Tarek Dghaily}, {and}
  \bibinfo{person}{Bernard Ghanem}.} \bibinfo{year}{2018}\natexlab{}.
\newblock \showarticletitle{{SoccerNet}: A Scalable Dataset for Action Spotting
  in Soccer Videos}. In \bibinfo{booktitle}{\emph{IEEE Int. Conf. Comput. Vis.
  and Pattern Recogn. Work. (CVPRW), CVsports}}. \bibinfo{publisher}{Inst.
  Elect. and Electron. Engineers (IEEE)}, \bibinfo{address}{Salt Lake City, UT,
  USA}, \bibinfo{pages}{1711--1721}.
\newblock
\urldef\tempurl%
\url{https://doi.org/10.1109/CVPRW.2018.00223}
\showDOI{\tempurl}


\bibitem[Hendrycks et~al\mbox{.}(2019)]%
        {Hendrycks2020AugMixAS}
\bibfield{author}{\bibinfo{person}{Dan Hendrycks}, \bibinfo{person}{Norman Mu},
  \bibinfo{person}{Ekin~D. Cubuk}, \bibinfo{person}{Barret Zoph},
  \bibinfo{person}{Justin Gilmer}, {and} \bibinfo{person}{Balaji
  Lakshminarayanan}.} \bibinfo{year}{2019}\natexlab{}.
\newblock \showarticletitle{{AugMix}: A Simple Data Processing Method to
  Improve Robustness and Uncertainty}.
\newblock \bibinfo{journal}{\emph{arXiv}}  \bibinfo{volume}{abs/1912.02781}
  (\bibinfo{year}{2019}), \bibinfo{numpages}{15}~pages.
\newblock
\showeprint{1912.02781}


\bibitem[Huang et~al\mbox{.}(2021)]%
        {huang2021pp}
\bibfield{author}{\bibinfo{person}{Xin Huang}, \bibinfo{person}{Xinxin Wang},
  \bibinfo{person}{Wenyu Lv}, \bibinfo{person}{Xiaying Bai},
  \bibinfo{person}{Xiang Long}, \bibinfo{person}{Kaipeng Deng},
  \bibinfo{person}{Qingqing Dang}, \bibinfo{person}{Shumin Han},
  \bibinfo{person}{Qiwen Liu}, \bibinfo{person}{Xiaoguang Hu}, {et~al\mbox{.}}}
  \bibinfo{year}{2021}\natexlab{}.
\newblock \showarticletitle{PP-YOLOv2: A practical object detector}.
\newblock \bibinfo{journal}{\emph{arXiv}}  \bibinfo{volume}{abs/2104.10419}
  (\bibinfo{year}{2021}), \bibinfo{numpages}{7}~pages.
\newblock
\showeprint{2104.10419}


\bibitem[Khosla et~al\mbox{.}(2020)]%
        {khosla2020supervised}
\bibfield{author}{\bibinfo{person}{Prannay Khosla}, \bibinfo{person}{Piotr
  Teterwak}, \bibinfo{person}{Chen Wang}, \bibinfo{person}{Aaron Sarna},
  \bibinfo{person}{Yonglong Tian}, \bibinfo{person}{Phillip Isola},
  \bibinfo{person}{Aaron Maschinot}, \bibinfo{person}{Ce Liu}, {and}
  \bibinfo{person}{Dilip Krishnan}.} \bibinfo{year}{2020}\natexlab{}.
\newblock \showarticletitle{Supervised Contrastive Learning}. In
  \bibinfo{booktitle}{\emph{Advances in Neural Information Processing
  Systems}}, \bibfield{editor}{\bibinfo{person}{H.~Larochelle},
  \bibinfo{person}{M.~Ranzato}, \bibinfo{person}{R.~Hadsell},
  \bibinfo{person}{M.F. Balcan}, {and} \bibinfo{person}{H.~Lin}} (Eds.),
  Vol.~\bibinfo{volume}{33}. \bibinfo{publisher}{Curran Associates, Inc.},
  \bibinfo{address}{Virtual conference}, \bibinfo{pages}{18661--18673}.
\newblock


\bibitem[Lin et~al\mbox{.}(2020)]%
        {Lin2020FocalLF}
\bibfield{author}{\bibinfo{person}{Tsung-Yi Lin}, \bibinfo{person}{Priya
  Goyal}, \bibinfo{person}{Ross Girshick}, \bibinfo{person}{Kaiming He}, {and}
  \bibinfo{person}{Piotr Doll{\'a}r}.} \bibinfo{year}{2020}\natexlab{}.
\newblock \showarticletitle{Focal Loss for Dense Object Detection}.
\newblock \bibinfo{journal}{\emph{IEEE Trans. Pattern Anal. Mach. Intell.}}
  \bibinfo{volume}{42}, \bibinfo{number}{2} (\bibinfo{date}{Feb.}
  \bibinfo{year}{2020}), \bibinfo{pages}{318--327}.
\newblock
\urldef\tempurl%
\url{https://doi.org/10.1109/tpami.2018.2858826}
\showDOI{\tempurl}


\bibitem[Liu et~al\mbox{.}(2021)]%
        {Liu2021VideoST}
\bibfield{author}{\bibinfo{person}{Ze Liu}, \bibinfo{person}{Jia Ning},
  \bibinfo{person}{Yue Cao}, \bibinfo{person}{Yixuan Wei},
  \bibinfo{person}{Zheng Zhang}, \bibinfo{person}{Stephen Lin}, {and}
  \bibinfo{person}{Han Hu}.} \bibinfo{year}{2021}\natexlab{}.
\newblock \showarticletitle{Video Swin Transformer}.
\newblock \bibinfo{journal}{\emph{arXiv}}  \bibinfo{volume}{abs/2106.13230}
  (\bibinfo{year}{2021}), \bibinfo{numpages}{12}~pages.
\newblock
\showeprint{2106.13230}


\bibitem[Luiten et~al\mbox{.}(2021)]%
        {Luiten2021Hota}
\bibfield{author}{\bibinfo{person}{Jonathon Luiten}, \bibinfo{person}{Aljosa
  Osep}, \bibinfo{person}{Patrick Dendorfer}, \bibinfo{person}{Philip Torr},
  \bibinfo{person}{Andreas Geiger}, \bibinfo{person}{Laura Leal-Taix{\'e}},
  {and} \bibinfo{person}{Bastian Leibe}.} \bibinfo{year}{2021}\natexlab{}.
\newblock \showarticletitle{{HOTA}: A higher order metric for evaluating
  multi-object tracking}.
\newblock \bibinfo{journal}{\emph{Int. J. Comput. Vis.}} \bibinfo{volume}{129},
  \bibinfo{number}{2} (\bibinfo{date}{Oct.} \bibinfo{year}{2021}),
  \bibinfo{pages}{548--578}.
\newblock
\urldef\tempurl%
\url{https://doi.org/10.1007/s11263-020-01375-2}
\showDOI{\tempurl}


\bibitem[Luo et~al\mbox{.}(2021a)]%
        {TransReid_SSL}
\bibfield{author}{\bibinfo{person}{Hao Luo}, \bibinfo{person}{Pichao Wang},
  \bibinfo{person}{Yi Xu}, \bibinfo{person}{Feng Ding}, \bibinfo{person}{Yanxin
  Zhou}, \bibinfo{person}{Fan Wang}, \bibinfo{person}{Hao Li}, {and}
  \bibinfo{person}{Rong Jin}.} \bibinfo{year}{2021}\natexlab{a}.
\newblock \showarticletitle{Self-Supervised Pre-Training for Transformer-Based
  Person Re-Identification}.
\newblock \bibinfo{journal}{\emph{arXiv}}  \bibinfo{volume}{abs/2111.12084}
  (\bibinfo{year}{2021}), \bibinfo{numpages}{15}~pages.
\newblock
\showeprint{2111.12084}


\bibitem[Luo et~al\mbox{.}(2021b)]%
        {Luo2021SelfSupervisedPF}
\bibfield{author}{\bibinfo{person}{Haowen Luo}, \bibinfo{person}{Pichao Wang},
  \bibinfo{person}{Yi Xu}, \bibinfo{person}{Feng Ding}, \bibinfo{person}{Yanxin
  Zhou}, \bibinfo{person}{Fan Wang}, \bibinfo{person}{Hao Li}, {and}
  \bibinfo{person}{Rong Jin}.} \bibinfo{year}{2021}\natexlab{b}.
\newblock \showarticletitle{Self-Supervised Pre-Training for Transformer-Based
  Person Re-Identification}.
\newblock \bibinfo{journal}{\emph{arXiv}}  \bibinfo{volume}{abs/2111.12084}
  (\bibinfo{year}{2021}), \bibinfo{numpages}{15}~pages.
\newblock
\showeprint{2111.12084}


\bibitem[Schroff et~al\mbox{.}(2015)]%
        {Schroff2015FaceNetAU}
\bibfield{author}{\bibinfo{person}{Florian Schroff}, \bibinfo{person}{Dmitry
  Kalenichenko}, {and} \bibinfo{person}{James Philbin}.}
  \bibinfo{year}{2015}\natexlab{}.
\newblock \showarticletitle{{FaceNet}: A unified embedding for face recognition
  and clustering}. In \bibinfo{booktitle}{\emph{IEEE/CVF Conf. Comput. Vis. and
  Pattern Recogn. (CVPR)}}. \bibinfo{publisher}{Inst. Elect. and Electron.
  Engineers (IEEE)}, \bibinfo{address}{Boston, MA, USA},
  \bibinfo{pages}{815--823}.
\newblock
\urldef\tempurl%
\url{https://doi.org/10.1109/cvpr.2015.7298682}
\showDOI{\tempurl}


\bibitem[Soares and Shah(2022)]%
        {soares2022action}
\bibfield{author}{\bibinfo{person}{Jo{\~a}o~V.~B. Soares} {and}
  \bibinfo{person}{Avijit Shah}.} \bibinfo{year}{2022}\natexlab{}.
\newblock \showarticletitle{Action Spotting using Dense Detection Anchors
  Revisited: Submission to the {SoccerNet} {Challenge} 2022}.
\newblock \bibinfo{journal}{\emph{arXiv}}  \bibinfo{volume}{abs/2206.07846}
  (\bibinfo{year}{2022}), \bibinfo{numpages}{3}~pages.
\newblock
\showeprint{2206.07846}


\bibitem[Soares et~al\mbox{.}(2022)]%
        {soares2022temporally}
\bibfield{author}{\bibinfo{person}{Jo{\~a}o~V.~B. Soares},
  \bibinfo{person}{Avijit Shah}, {and} \bibinfo{person}{Topojoy Biswas}.}
  \bibinfo{year}{2022}\natexlab{}.
\newblock \showarticletitle{Temporally Precise Action Spotting in Soccer Videos
  Using Dense Detection Anchors}.
\newblock \bibinfo{journal}{\emph{arXiv}}  \bibinfo{volume}{abs/2205.10450}
  (\bibinfo{year}{2022}), \bibinfo{numpages}{5}~pages.
\newblock
\showeprint{2205.10450}


\bibitem[van~den Oord et~al\mbox{.}(2018)]%
        {Oord2018RepresentationLW}
\bibfield{author}{\bibinfo{person}{A{\"a}ron van~den Oord},
  \bibinfo{person}{Yazhe Li}, {and} \bibinfo{person}{Oriol Vinyals}.}
  \bibinfo{year}{2018}\natexlab{}.
\newblock \showarticletitle{Representation Learning with Contrastive Predictive
  Coding}.
\newblock \bibinfo{journal}{\emph{arXiv}}  \bibinfo{volume}{abs/1807.03748}
  (\bibinfo{year}{2018}), \bibinfo{numpages}{13}~pages.
\newblock
\showeprint{1807.03748}


\bibitem[Wang et~al\mbox{.}(2007)]%
        {Wang2007ShapeAA}
\bibfield{author}{\bibinfo{person}{Xiaogang Wang}, \bibinfo{person}{Gianfranco
  Doretto}, \bibinfo{person}{Thomas Sebastian}, \bibinfo{person}{Jens
  Rittscher}, {and} \bibinfo{person}{Peter Tu}.}
  \bibinfo{year}{2007}\natexlab{}.
\newblock \showarticletitle{Shape and Appearance Context Modeling}. In
  \bibinfo{booktitle}{\emph{Int. Conf. Comput. Vis.}} \bibinfo{publisher}{Inst.
  Elect. and Electron. Engineers (IEEE)}, \bibinfo{address}{Rio de Janeiro,
  Brazil}, \bibinfo{pages}{1--8}.
\newblock
\urldef\tempurl%
\url{https://doi.org/10.1109/iccv.2007.4409019}
\showDOI{\tempurl}


\bibitem[Wortsman et~al\mbox{.}(2022)]%
        {Wortsman2022ModelSA}
\bibfield{author}{\bibinfo{person}{Mitchell Wortsman}, \bibinfo{person}{Gabriel
  Ilharco}, \bibinfo{person}{Samir~Yitzhak Gadre}, \bibinfo{person}{Rebecca
  Roelofs}, \bibinfo{person}{Raphael Gontijo-Lopes}, \bibinfo{person}{Ari~S.
  Morcos}, \bibinfo{person}{Hongseok Namkoong}, \bibinfo{person}{Ali Farhadi},
  \bibinfo{person}{Yair Carmon}, \bibinfo{person}{Simon Kornblith}, {and}
  \bibinfo{person}{Ludwig Schmidt}.} \bibinfo{year}{2022}\natexlab{}.
\newblock \showarticletitle{Model soups: averaging weights of multiple
  fine-tuned models improves accuracy without increasing inference time}.
\newblock \bibinfo{journal}{\emph{arXiv}}  \bibinfo{volume}{abs/2203.05482}
  (\bibinfo{year}{2022}), \bibinfo{numpages}{34}~pages.
\newblock
\showeprint{2203.05482}


\bibitem[Ye et~al\mbox{.}(2022)]%
        {Ye2022DeepLF}
\bibfield{author}{\bibinfo{person}{Mang Ye}, \bibinfo{person}{Jianbing Shen},
  \bibinfo{person}{Gaojie Lin}, \bibinfo{person}{Tao Xiang},
  \bibinfo{person}{Ling Shao}, {and} \bibinfo{person}{Steven C.~H. Hoi}.}
  \bibinfo{year}{2022}\natexlab{}.
\newblock \showarticletitle{Deep Learning for Person Re-Identification: A
  Survey and Outlook}.
\newblock \bibinfo{journal}{\emph{IEEE Trans. Pattern Anal. Mach. Intell.}}
  \bibinfo{volume}{44}, \bibinfo{number}{6} (\bibinfo{date}{June}
  \bibinfo{year}{2022}), \bibinfo{pages}{2872--2893}.
\newblock
\urldef\tempurl%
\url{https://doi.org/10.1109/tpami.2021.3054775}
\showDOI{\tempurl}


\bibitem[Zheng et~al\mbox{.}(2015)]%
        {Zheng2015ScalablePR}
\bibfield{author}{\bibinfo{person}{Liang Zheng}, \bibinfo{person}{Liyue Shen},
  \bibinfo{person}{Lu Tian}, \bibinfo{person}{Shengjin Wang},
  \bibinfo{person}{Jingdong Wang}, {and} \bibinfo{person}{Qi Tian}.}
  \bibinfo{year}{2015}\natexlab{}.
\newblock \showarticletitle{Scalable Person Re-identification: A Benchmark}. In
  \bibinfo{booktitle}{\emph{Int. Conf. Comput. Vis.}} \bibinfo{publisher}{Inst.
  Elect. and Electron. Engineers (IEEE)}, \bibinfo{address}{Santiago, Chile},
  \bibinfo{pages}{1116--1124}.
\newblock
\urldef\tempurl%
\url{https://doi.org/10.1109/iccv.2015.133}
\showDOI{\tempurl}


\bibitem[Zhong et~al\mbox{.}(2020)]%
        {Zhong2020RandomED}
\bibfield{author}{\bibinfo{person}{Zhun Zhong}, \bibinfo{person}{Liang Zheng},
  \bibinfo{person}{Guoliang Kang}, \bibinfo{person}{Shaozi Li}, {and}
  \bibinfo{person}{Yi Yang}.} \bibinfo{year}{2020}\natexlab{}.
\newblock \showarticletitle{Random Erasing Data Augmentation}. In
  \bibinfo{booktitle}{\emph{AAAI}}, Vol.~\bibinfo{volume}{34}.
  \bibinfo{publisher}{Association for the Advancement of Artificial
  Intelligence}, \bibinfo{address}{New York, USA},
  \bibinfo{pages}{13001--13008}.
\newblock
\urldef\tempurl%
\url{https://doi.org/10.1609/aaai.v34i07.7000}
\showDOI{\tempurl}


\bibitem[Zhou et~al\mbox{.}(2021)]%
        {Zhou2021FeatureCM}
\bibfield{author}{\bibinfo{person}{Xin Zhou}, \bibinfo{person}{Le Kang},
  \bibinfo{person}{Zhiyu Cheng}, \bibinfo{person}{Bo He}, {and}
  \bibinfo{person}{Jingyu Xin}.} \bibinfo{year}{2021}\natexlab{}.
\newblock \showarticletitle{Feature Combination Meets Attention: Baidu Soccer
  Embeddings and Transformer based Temporal Detection}.
\newblock \bibinfo{journal}{\emph{arXiv}}  \bibinfo{volume}{abs/2106.14447}
  (\bibinfo{year}{2021}), \bibinfo{numpages}{7}~pages.
\newblock
\showeprint{2106.14447}


\end{thebibliography}

\newpage
\appendix
\section{Appendix}
\label{sec:appendix}

In this appendix, the participants provide a short summary of their methods. Only teams who provided a technical report at the end of the challenge that has been peer-reviewed by the organizers were able to submit a summary. This is to ensure that the presented methods followed the challenge rules.

\subsection{Action Spotting} \label{app:spotting}
The complete leaderboard is provided in Table \ref{tab:LeaderboardSpotting}.

\mysection{S3 - Temporal Action Detection}\\
\textit{Wei Li, Shimin Chen, Jianyang Gu, Chen Chen, Ning Wang, and Yandong Guo
}\\
\textit{liwei19@oppo.com, chenshimin1@oppo.com, \\gu\_jianyang@zju.edu.cn, chenchen@oppo.com,  \\wangning12@mail.ecust.edu.cn, guoyandong@oppo.com}

We apply temporal action detection (TAD) method following Faster-TAD~\cite{chen2022faster} to the action spotting task. We use a 16 seconds sliding window with 8 seconds stride to detect a 4 seconds soccer action. The starting position of an action is the officially provided timestamp. We first generate multiple VideoSwinTransformer~\cite{Liu2021VideoST} feature extractors pre-trained on different customized label styles which are designed to be sensitive to action boundary. Then, we generate temporal proposals and semantic labels in a unified network following Faster-TAD, learning useful context for each proposal. We involve cross entropy loss and triplet loss~\cite{Schroff2015FaceNetAU} for explicit constraints of embedded feature distributions. After collecting all TAD results for each sliding window, we use NMS to generate final results of a game. With some ensemble methods, we finally reached 68.46\% and 64.88\% tight Average-mAP on SoccerNet test and challenge sets, respectively.

\mysection{S4 - Group-wise Multi-scale action Detector (GMDet)}\\
\textit{Zhiheng Li, Wei Zhang, Yujie Zhong, Dengjie Li, Feng Yan, Xiaolin Wei, Ran Song, and Lin Ma}\\
\textit{zhihengli@mail.sdu.edu.cn, davidzhang@sdu.edu.cn, \\zhongyujie@meituan.com, lidengjie@meituan.com, \\yanfeng05@meituan.com, weixiaolin02@meituan.com, \\ransong@sdu.edu.cn, linma@alumni.cuhk.net}

The Group-wise Multi-scale action Detector (GMDet) aims at detecting/spotting actions in untrimmed videos. It consists of an ensembled feature encoder (proposed by Baidu, the 1st place last year), a multi-scale transformer encoder and a CNN-based decoder (which predicts the action classes and temporal locations). In order to handle various actions with very different properties, GMDet divides actions into three action groups, where similar actions belong to the same group. The group-wise operation enables a specific network architecture and a tailored training strategy for each group. First, we adopt decoders of different depth for the three action groups. Second, Mix-up and adaptive loss weights are leveraged for the action group with few training examples. At inference, one or more models (in a way of model ensemble) are used to predict action instances for each group independently. The proposed GMDet demonstrates superior performance on the action spotting task.

\begin{table}[h]
    \caption{Action spotting leaderboard. Main metric for the leaderboard and best performances in bold. Team names with a superscript have provided a summary that may be found in Appendix~\ref{app:spotting} or in Section~\ref{sub:spottingwinner} for the winning team.
    The baseline description may be found in \url{https://github.com/SoccerNet/sn-spotting}.
    }
    \label{tab:LeaderboardSpotting}
    \centering
\resizebox{\linewidth}{!}{
    \begin{tabular}{l||c|c|c||c|c|c}
\multirow{ 2}{*}{Participants} & \multicolumn{3}{c||}{\textbf{tight Average-mAP}} & \multicolumn{3}{c}{loose Average-mAP} \\ \cline{2-7}
& \textbf{main} & vis. & inv. & main & vis. & inv. \\ \midrule
\bf{Yahoo Research$^{S1}$} & \bf{67.81} & 72.84  & 60.17 & 78.05  & 80.61 & 78.05 \\
PTS & 66.73 & 74.84  & 53.21 & 73.62  & 79.16 & 67.42 \\
AS\&RG$^{S3}$ & 64.88 & 70.31  & 53.03 & 72.83  & 76.08 & 72.35 \\
mt\_sdu\_action$^{S4}$ & 62.26 & 67.48  & 45.04 & 69.86  & 73.81 & 59.15\\
Rkrystal & 61.84 & 67.39  & 48.71 & 74.75  & 78.29 & 69.02 \\ 
arturxe$^{S6}$ & 60.56 & 65.75 & 53.00 & 71.72  & 75.15 & 69.91 \\
cihe$^{S7}$ & 59.97 & 64.51  & 53.80 & 72.95  & 76.29 & 71.95 \\
GUC$^{S8}$ & 58.71 & 63.70  & 51.86 & 70.49  & 73.46 & 70.11 \\
abcdefg & 56.07 & 62.97  & 46.51 & 67.88  & 72.54 & 66.37 \\
intro- and inter & 	53.97 & 60.04  & 47.52 & 67.75  & 71.16 & 70.12 \\
memory & 53.03 & 57.94 & 43.16 & 67.15  & 69.20 & 68.28 \\
stargazer$^{S12}$ & 52.04 & 60.18  & 32.06 & 60.86  & 66.64 & 48.46 \\
heaven & 	51.85 & 59.85  & 31.62 & 60.88  & 66.67 & 48.45 \\
lczazu & 49.56 & 56.82 & 31.60 & 60.86 & 66.56 & 48.51 \\
Baseline* & 49.56* & 54.42  & 45.42 & 74.84  & 78.58 & 71.52 \\
zqing & 47.54 & 51.75 & 41.65 & 66.66 & 69.06 & 67.17 \\
welkin & 42.74 & 49.91  & 20.67 & 50.90 & 56.48 & 35.38 \\
DUT & 40.65 & 43.87 & 43.10 & 68.40 & 71.68 & 68.53 \\
sshinde5 & 36.71 & 39.33 & 21.26 & 51.36 & 55.29 & 35.34 \\
SIT$^{S20}$ & 21.60 & 26.55 & 16.83 & 29.92 & 34.92 & 25.22 \\

    \end{tabular}
}
\end{table}

\mysection{S6 - Hierarchical Multimodal Transformers for Action Spotting (HMTAS)}\\
\textit{Artur Xarles, Sergio Escalera, Albert Clapés
Thomas B. Moeslund, and Rikke Gade}\\
\textit{arturxe@gmail.com, sergio.escalera.guerrero@gmail.com,\\ alcl@create.aau.dk, tbm@create.aau.dk, rg@create.aau.dk}

Hierarchical Multimodal Transformers for Action Spotting (HMTAS) is built on 6 different precomputed embeddings, the 5 visual ones from TPN, GTA, VTN, irCSN, and I3D-Slow backbones (provided by the winners of the action spotting challenge at CVPR 2021 ActivityNet workshop \cite{Zhou2021FeatureCM}), plus the audio ones from a VGGish fine-tuned on SoccerNet-v2. Differently from \cite{Zhou2021FeatureCM}, which concatenated the visual embeddings channel-wise, the sequences of the different embeddings accounting for 3-seconds-long temporal windows are here independently evolved through their own Transformer encoder. The evolved embeddings are then temporally max-pooled and concatenated into a 5+1 sequence of global embeddings that is fed to a multimodal Transformer encoder producing the final multilabel classification into the 18 different spotting classes. HMTAS minimizes the Negative Log-Likelihood Loss (NLLL) of the final classification, but also the auxiliary NLLLs supervising the intermediate classifications attempted by the unimodal encoders. During inference, Non-Maximum Suppression is used to reduce the number of spotting candidates.

\mysection{S7 - Mixed Spatial and Temporal Attention for Soccer Game Action Spotting (CIHE)}\\
\textit{Cheuk-Yiu Chan, Chun-Chuen Hui, Wan-Chi Siu, Sin-wai Chan, and H. Anthony Chan}\\
\textit{cy3chan@cihe.edu.hk, cchui@cihe.edu.hk, enwcsiu@polyu.edu.hk,\\chansinwai@cihe.edu.hk, hhchan@cihe.edu.hk}

A novel approach on mixing spatial and temporal attention is proposed. Our model consists of two
similar stages, with the 2nd stage designed for improving spatio-temporal representation capability by
enlarging receptive fields. Each stage has initially a Transformer network (Temporal-Grouped
Attention (TGA)) with different sets of parameters to be trained for different groups of channels, and
another transformer with only one set of parameters formed by grouping all channels together for the
advantage of extracting temporal domain features. Each transformer contains a concurrent “Temporal-
Grouped Local Attention” network and a “Temporal-Grouped Global Attention’ network. A Selective
Feature Aggregation (SFA) structure is proposed to select the final weights between the two attention
networks intelligently which forms an additional attention-based gate controlling mechanism, and
allows a further cooperation of the spatial and temporal features for making good decisions on soccer
game action spotting. Our network has achieved a tight Average mAP of 60.51\%.
\textit{Acknowledgement:} Work supported by Hong Kong UGC Grant: UGC/IDS(C)11/E01/20) via CIHE.

\mysection{S8 - STE}\\
\textit{Abdulrahman Darwish and Tallal El-Shabrawy}\\
\textit{abdulrahman.darwish@guc.edu.eg, tallal.el-shabrawy@guc.edu.eg}

STE is a proposed deep model for action spotting in soccer videos. The model reads the pre-extracted Baidu soccer embeddings, generated from SoccerNet-v2 dataset. It encodes the spatial features then the temporal features at different scales. Its architecture consists of 3 blocks: the spatial encoder, the temporal encoder then the prediction block. The first block extracts the spatial semantics through 1D max-pooling, a fully connected layer and 1D convolutional layer. The temporal encoder then extracts features across frames by applying 3 1D convolutional layers with 1D max-pooling after the first layer. The last block predicts two outputs, the first prediction is the class event that occurred in the input window. Two fully connected layers are used to map the extracted semantics to predicted class. The second output is the time frame where the event occurred. In a parallel branch, 2 fully connected layers predict the timestamp of the event.

\mysection{S12 - Stargazer}\\
\textit{He Zhu, Junwei Liang, Chengzhi Lin, and Jun Zhang, Jianming Hu}\\
\textit{zhuh20@mails.tsinghua.edu.cn, junweiliang1114@gmail.com, \\linchzh3@mail2.sysu.edu.cn, bobbyjzhang@tencent.com, \\hujm@mail.tsinghua.edu.cn}

We propose Stargazer, a transformer-based system which can efficiently exploit the rich temporal features about the soccer action information. We first extract a certain number of sequential frames from the video as clips, and then put them into the action recognition module pretrained on Kinetics. It is an improved network based on multi-scale vision transformer, and it can learn a hierarchy of robust representations. It processes the proposal clips and classifies them into one of the soccer actions or background. We also use spatial and temporal multi-crop data augmentation to facilitate the training. Finally, we utilize the action recognition module and overlapped sliding window strategy to extract the features of the video and sent them to model of NetVLAD++ to get the result.

\mysection{S20 - Simultaneous Supervised Contrastive Learning for Spatio-Temporal Action Recognition}\\
\textit{Andrei Boiarov and Nikita Kasatkin}\\
\textit{andrei.boiarov@sit.team, nk@sit.team}

Our approach is a combination of a spatio-temporal backbone and metric learning for the action recognition problem. The model consists of two networks: backbone (frozen pretrined on Kinetics-400 SlowFast-R101~\cite{feichtenhofer2019slowfast}) and head.
Head network: $L_2$ normalization, two fully connected (FC) blocks (Dropout, FC layer with hidden size $2048$, BatchNorm and ELU activation), embedding layer (output size $512$) and classification FC layer.
Dataset videos have been split as clips of 2 seconds duration.
We propose Simultaneous Supervised Contrastive~(Equation \ref{eq:simsupcon}) loss  for simultaneous training embedding and classifier in a~\cite{boiarov2019large} fashion. In contrast to the standard SupCon~(Equation \ref{eq:supcon})~\cite{khosla2020supervised}, function~(Equation  \ref{eq:simsupcon}) is calculated only for positive classes.

\begin{equation}\label{eq:ce}
    \mathcal{L}_{CE} = - \sum_{i \in I} \log \frac{\exp (W^{\intercal}_{\mathbf{y}_i \mathbf{z}_i})}{\sum_{j=1}^{n+1} \exp (W^\intercal_j \mathbf{z}_i)}
\end{equation}

\begin{equation}\label{eq:supcon}
    \mathcal{L}_{SC} = \sum_{\substack{i \in I\\ \mathbf{y}_i < n+1}} \frac{-1}{|P(i)|} \sum_{p \in P(i)} \log \frac{\exp(\mathbf{z}_i \mathbf{z}_p / \tau) }{\sum_{a \in A(i)} \exp(\mathbf{z}_i \mathbf{z}_a / \tau)}
\end{equation}

\begin{equation}\label{eq:simsupcon}
    \mathcal{L} = \mathcal{L}_{CE} + \frac{\lambda}{2} \mathcal{L}_{SC},
\end{equation}
$n$ --- positive classes, $n+1$ --- negative class, $I$ --- batch, $A(i)= I$ \textbackslash $\{i\}$, $P(i)$ --- one class indices, $\mathbf{z}_i$ --- $i$th embedding, $\mathbf{y}_i$ --- correct label, $W_j$ --- $j$th column of the classifier weights matrix, $\tau$ --- temperature.





\subsection{Field Localization}
The complete leaderboard is provided in Table \ref{tab:LeaderboardFieldLocalization}.

\label{app:pitch}
\begin{table}[t]
\caption{Field localization leaderboard. Main metric for the leaderboard and best performance in bold. Team names with a superscript provided a summary that can be found in Appendix \ref{app:pitch}, or in Section \ref{sub:fieldlocwinner} for the winner.\label{tab:LeaderboardFieldLocalization}
The baseline description may be found in \url{https://github.com/SoccerNet/sn-calibration}.}

\begin{tabular}{l||c|c|c||c}
Participants & \AE@5 & \AE@10 & \AE@20 & \textbf{Final score}\tabularnewline
\midrule 
ONEDAY$^{P1}$ & 84.40 & 90.24 & 92.17 & \textbf{87.61}\tabularnewline
\hline 
imgo$^{P2}$ & 74.19 & 84.59 & 87.62 & 79.84\tabularnewline
\hline 
2Ai-IPCA$^{P3}$ & 71.01 & 76.18 & 77.60 & 73.81\tabularnewline
\hline 
channings & 66.87 & 78.19 & 81.65 & 73.05\tabularnewline
\hline 
test222 & 61.8 & 77.42 & 81.45 & 70.21\tabularnewline
\hline 
eidos.ai & 63.42 & 75.94 & 78.32 & 70.04\tabularnewline
\hline 
Mike Azatov$^{P7}$ & 62.09 & 73.47 & 76.79 & 68.28\tabularnewline
\hline 
goahead & 50.67 & 82.87 & 92.54 & 68.28\tabularnewline
\hline 
L3S & 20.57 & 45.44 & 64.38 & 35.85\tabularnewline
\hline 
test26 & 15.91 & 44.02 & 61.29 & 32.56\tabularnewline
\hline 
tactica & 15.90 & 44.0 & 61.24 & 32.54\tabularnewline
\hline 
B1 & 14.65 & 40.41 & 56.47 & 29.94\tabularnewline
\hline 
Baseline* & 13.32 & 38.28 & 53.87 & 28.14*\tabularnewline
\end{tabular}

\end{table}

\mysection{P2 - MGTV}\\
\textit{Lingchi Chen, Yi Yu, Xinying Wang, and Jin Chen}\\
\textit{lingchi@mgtv.com, yuyi@mgtv.com, \\xinying@mgtv.com, chenjing@mgtv.com}

This solution comes from MGTV, we are a video media company from China. There is three main parts to introduce our method:
firstly, We modify the LaneNet network structure for semantic segmentation, which include a segmentation branch and instance branch. 
The second is About data augmentation, we have tried some enhancement methods, and the experiments show that the following three types of operations are most helpful for the prediction effect: left and right mirroring, randomcrop, and force the label of the image behind the goal to the left. The above three operations have brought about an improvement of nearly 8\% in the \AE@5 score.
Finally, the postprocessing of taking endpoints from line segments, also includes operations such as extending, correcting categories, and merging line under the guidance of the soccerpitch template. The line correction operations have brought about an improvement of nearly 12\% in the \AE@5 score.

\mysection{P3 - Hierarchical Line Extremity Segmentation Unet}\\
\textit{José Henrique Brito and Miguel Santos Marques
}\\
\textit{jbrito@ipca.pt, a18888@alunos.ipca.pt}

Our method directly infers line extremities using a CNN based on Unet. The input and outputs are $288\times512$. It produces three outputs, the first with 3 channels (Background, Floor Lines, Goal Lines), the second 26 channels (line classes), the third 25 channels, (class extremities except “Circle central”). Output 1 has a Softmax activation, and the others use Sigmoid. Outputs are hierarchical, \ie, output 2 takes as input the concatenation of the last backbone feature map and the output 1 feature map. Output 3 takes as input the concatenation of the last backbone feature map and the output 2 feature map. The global loss is a weighted average the 3 output entropy losses. Line masks for training were simply drawn, extremity masks were generated with Gaussian heatmaps centered on extremity coordinates with standard deviation of 1. Inferred extremity masks are thresholded and the highest pair of peaks are used as predictions.

\mysection{P7 - Two-step procedure}\\
\textit{Mike Azatov}\\
\textit{mazatov@gmail.com}

The approach follows a two-step procedure where in the first step I find the pitch element segmentation of the image and in the second part, I find the locations of the extremities. The segmentation is improved by using a superior segmentation model coupled with splitting the dataset into a few sub-datasets based on the camera view. Multiple camera views challenge segmentation algorithms as different pitch elements can appear in different positions with respect to each other depending on the view. The extremity localization is done by fitting lines and ellipses to individual pitch elements and finding their intersection points with each other as well as with the image borders.

\subsection{Camera Calibration}
\label{app:camera}
The complete leaderboard is provided in Table \ref{tab:LeaderboardCalibration}.

\begin{table}[t]
\caption{Camera calibration leaderboard. Main metric for the leaderboard and
best performance in bold. Team names with a superscript provided a
summary that can be found in Appendix \ref{app:camera},
or in Section \ref{sub:camerawinner} for
the winner.\label{tab:LeaderboardCalibration} The baseline description may be found in \url{https://github.com/SoccerNet/sn-calibration}.}

\begin{tabular}{l||c|c|c|c||c}

Participants & \AS@5 & \AS@10 & \AS@20 & \CR & \textbf{Final $s$}\tabularnewline
 
\midrule 
achengmao$^{C1}$ & 82.38 & 94.80 & 96.33 & 72.61 & \textbf{83.96}\tabularnewline
\hline 
L3S$^{C2}$ & 57.83 & 81.42 & 90.74 & 69.32 & 66.58\tabularnewline
\hline 
MikeAzatov$^{C3}$ & 62.25 & 84.32 & 90.56 & 56.41 & 66.45\tabularnewline
\hline 
2AI-IPCA$^{C4}$ & 51.31 & 65.47 & 71.43 & 60.57 & 54.03\tabularnewline
\hline 
imgo & 45.11 & 61.38 & 68.63 & 77.98 & 51.93\tabularnewline
\hline 
test26 & 13.05 & 28.49 & 41.66 & 68.90 & 21.30*\tabularnewline
\hline 
Baseline* & 12.94 & 29.14 & 43.48 & 58.95 & 21.00*\tabularnewline
\end{tabular}

\end{table}

\mysection{C2 - Camera Calibration for Broadcast Soccer Videos}\\
\textit{Jonas Theiner and Ralph Ewerth}\\
\textit{theiner@l3s.de, ralph.ewerth@tib.eu}

We propose the differentiable \emph{segment reprojection loss} function that aims to approximate the 6-DoF camera pose~(position and orientation) and focal length given segment correspondences~(lines and point clouds) between image and calibration object~(3D~soccer pitch model). 
Initialized with camera parameters representing \emph{central} views from multiple camera locations, the \emph{segment reprojection loss} induced by the camera parameters is iteratively minimized via gradient descent.
Estimates that are likely invalid~(\eg, erroneous pitch element localization, local minima, out of camera distribution) are automatically discarded by thresholding the loss. 
For the segmentation of pitch elements and subsequent selection of relevant pixels, we apply a retrained baseline segmentation model and randomly sample additional points to get more stable gradients.

\mysection{C3 - Self-Assessing Algorithm}\\
\textit{Mike Azatov}\\
\textit{mazatov@gmail.com}

The core idea in the camera calibration algorithm is to combine the location of extremities with pitch segmentation results to create a self-assessing algorithm that can accurately predict how well the camera is calibrated and decide on whether further improvements are necessary. The point and line correspondences provided a candidate solution for camera parameters. The self-assessment mechanism decides whether the results are sufficiently strong, and proceeds to optimize the camera calibration until it we get satisfactory results

\mysection{C4 - 2Ai-IPCA}\\
\textit{José Henrique Brito and Miguel Santos Marques}\\
\textit{jbrito@ipca.pt and a18888@alunos.ipca.pt}

Our camera calibration method uses the line extremity estimates produced by our pitch localization method and feeds those point coordinates as input to the baseline method for camera calibration provided by the SoccerNet team. The baseline camera calibration method then computes the homography between the detected 2D markings on the image and the known 3D measurements of the markings in a soccer field, and subsequently computes the camera calibration parameters.

\subsection{Player Re-Identification}
The complete leaderboard is provided in Table \ref{tab:LeaderboardReID}.
\label{app:reid}

\begin{table}[t]
\caption{Re-identification leaderboard. Main metric for the leaderboard and best performance in bold. Team names with a superscript have provided a summary that can be found in Appendix \ref{app:reid}, or in Section~\ref{sub:reidwinner} for the winner. The baseline description may be found in \url{https://github.com/SoccerNet/sn-reid}.}
\label{tab:LeaderboardReID}
\centering
\begin{tabular}{l||c|c@{\hspace*{0.3cm}}|@{\hspace*{0.08cm}}|@{\hspace*{0.3cm}}l||c|c}
Participants & \textbf{mAP} & R-1 & Participants & \textbf{mAP} & R-1 \\ \midrule
\textbf{Inspur$^{R1}$}  & \textbf{91.68} & 89.41 & tianchao        & 87.39 & 84.25\\
MGSoccer$^{R2}$         & 91.48 & 89.21 & 126\_187        & 86.96 & 83.04\\
MTVACV$^{R3}$           & 90.11 & 87.04 & Blackghost      & 78.25 & 71.50\\
gunners$^{R4}$          & 89.47 & 86.42 & 1234567         & 65.98 & 58.40\\
MIG$^{R5}$              & 89.44 & 86.11 & rasengan        & 64.79 & 54.27\\
MMLab                   & 89.01 & 85.68 & baba            & 60.32 & 48.91\\
ReBatch$^{R7}$          & 88.36 & 84.82 & Baseline*        & 59.11* & 48.41\\
\end{tabular}

\end{table}


\mysection{R2 - MGSoccer-ViT}\\
\textit{Guanshuo Wang, Junjie Li, Fufu Yu, Qiong Jia, Shouhong Ding}\\
\textit{mediswang@tencent.com, serenitycapo@gmail.com, \\fufuyu@tencent.com, boajia@tencent.com, \\ericshding@tencent.com}

We apply a hybrid architecture named MGSoccer-ViT to solve the player re-identification task. The player image is respectively represented by CNN and ViT backbones. CNN backbone is designed according to the Multiple Granularity Network, which represents global and local features by independent branches with different levels of partitions. ViT backbone is referred to the TransReID architecture, which divides the images into equal patches as input token sequences of Transformer networks. All the backbones are initialized with self-supervised pretrained models on large-scale LUPerson dataset. ArcFace and hard triplet losses are applied during training. Towards the special data distribution of SoccerNet, we remove all the long-tail PIDs and apply an action-based sampler to sample equal numbers of PIDs in different actions. Various data augmentation methods are employed including random flipping, erasing, cropping after padding, AutoAugment and AugMix. Our model achieves 91.28\% mAP on the Test set, and 91.48\% on the Challenge set.

\mysection{R3 - Domain-Aware Self-Supervised Pre-Training}\\
\textit{Siyu Chen, Dengjie Li, Yujie Zhong, Fan Liang, Xiaolin Wei, Lin Ma}\\
\textit{chensiyu25@meituan.com, lidengjie@meituan.com, \\zhongyujie@meituan.com liangfan02@meituan.com, \\weixiaolin02@meituan.com, forest.linma@gmail.com}

The SoccerNet ReID dataset mainly includes two difficulties: high appearance similarity of players and serious mutual occlusions between players. To alleviate these two problems, we first propose a target-domain-oriented self-supervised pre-training strategy to obtain a strong pretrained model. Namely, we refine the images of LUPerson dataset and retain those having high Catastrophic Forgetting Score (CFS) with the SoccerNet ReID dataset to form the pre-training set. DINO is adopted to train a ViT-B on the pre-training set. Second, in order to better distinguish different players, we design a visibility-and-importance-aware feature learning framework to extract discriminative features of different body parts of the players. Finally, we design a model fusion strategy, which simply averages the distance matrices of different models, to further boost the performance. Our method achieves 90.1 mAP on the SoccerNet ReID challenge dataset.

\mysection{R4 - Hierarchical sampling and centroid loss to improve player re-identification}\\
\textit{Bharath Comandur}\\
\textit{cjrbharath@gmail.com}

Instead of random sampling, we use a novel procedure to select samples for each batch. We use a loss function with centroids to better separate the embeddings. Finally we create a model-soup~\cite{Wortsman2022ModelSA} to further increase mAP. Our models are pretrained on only ImageNet. With these simple ideas, we achieve an mAP of 89.47 on the SoccerNet challenge set. More details can be found at~\cite{Comandur2022SportsRI}.

\mysection{R5 - MIG}\\
\textit{Leqi Shen and Tao He}\\
\textit{lunarshen@gmail.com, kevin.92.he@gmail.com}

We propose an improvement of the classical sampling strategy. We constitute batches by randomly sampling A actions, I identities and K images per identity, thus resulting in a batch of A × I × K images. The large number of overall actions and the randomness can prevent different identities of the same player from appearing at the same batch. We utilize non-parametric InfoNCE loss within batches. We use ViT with ICS in TransReID-SSL~\cite{Luo2021SelfSupervisedPF} as our backbone. Based on self-attention in Transformer, a features fusion method is proposed to generate the refined feature for an original query feature. Model ensemble is also an effective method to boost performance. Five models (different input sizes and pretrain models) with the fusion method are ensembled to generate the final distance matrix.

\mysection{R7 - Person re-identification task with RESNET ensemble}\\
\textit{Jorge De Corte, Andreas Luyts, and Ruben Debien}\\
\textit{jorge.decorte@rebatch.be, andreas.luyts@rebatch.be, \\ruben.debien@rebatch.be}

RE(s)IDnet is an ensemble of models with RESNET variants as backbone. These individual models are trained using a weighted sum of the Euclidian Triplet loss and an ID/classification loss, the latter only having a very small contribution to the overall loss. Images from the same action ID were kept together in batches to ensure that the best/hardest triplets possible were used. A cyclical learning rate schedule, starting with a warmup cycle, was used during the training process. Pre- and postprocessing have a big impact on model performance. Various augmentation techniques, such as coarse dropout, are used in the preprocessing phase. Postprocessing involves changing the gallery rankings after prediction. If all models agree on the highest ranked image for a certain query, then we can move this image backwards in the ranking for all other queries.

\subsection{Multiple Player Tracking}
The complete leaderboard is provided in Table \ref{tab:LeaderboardTracking}.
\label{app:tracking}
\begin{table}[t]
    \caption{Tracking leaderboard. Main metric for the leaderboard and best performances in bold. Team names with a superscript have provided a summary that may be found in Appendix~\ref{app:tracking}, or in Section~\ref{sub:trackingwinner} for the winning team. The baseline description may be found in \url{https://github.com/SoccerNet/sn-tracking}.}
    \label{tab:LeaderboardTracking}
    
    \centering
    \begin{tabular}{l||c|c|c}
Participants & \textbf{HOTA} & DetA & AssA \\ \midrule
\bf{Kalisteo$^{T1}$} & \bf{93.64} & 99.56 & 88.06 \\
CBIOUT (CB-IoU)$^{T2}$ & 93.25 & 99.76 & 87.15 \\
tactica$^{T3}$ & 93.17 & 99.85 & 86.94 \\
FGV & 92.49 & 99.76 &	85.74\\
smot & 91.49 & 99.77 & 83.90 \\
tianchao & 89.42 & 99.62 & 	80.27 \\
who & 88.99 & 99.74 & 79.39 \\
tomo & 88.94 & 99.77 & 79.28 \\
dk$^{T9}$ & 88.65 & 99.70 & 78.82 \\
1p & 88.55 & 99.68 &  78.67 \\
Baseline* & 	70.89* & 82.97 & 60.68 \\
ret-1 & 57.81 & 70.07 & 47.89 \\
WOTAICAILE & 51.03 & 60.83 & 42.96
    \end{tabular}
\end{table}

\mysection{T2 - Cascaded Buffered IoU (BIoU)}\\
\textit{Fan Yang, Shigeyuki Odashima, Shoichi Masui, and Shan Jiang}\\
\textit{fan.yang@fujitsu.com, sodashima@fujitsu.com, \\masui.shoichi@fujitsu.com, jiang.shan@fujitsu.com}

Our method mainly includes two steps. First, we propose a Cascaded Buffered IoU (BIoU) for online short-term tracking. Compared with normal IoU matching, buffers are added to observations in BIoU to relax the matching space, which could alleviate the miss matching caused by motion estimation errors in complicated scenes. Meanwhile, we apply cascaded matching to match simple ones before the hard ones. Then, we link short-term tracklets to long-term ones using Hierarchical Clustering (HC). The distance matrix used in HC is formed by comparing appearance features extracted from short-term tracklets.

\mysection{T3 - Camera-motion-aware appearance-based ByteTrack}\\
\textit{Fang Da}\\
\textit{fang@qcraft.ai}

Our submission entry builds on the ByteTrack baseline. We augment the track-detection similarity metric with a track appearance feature extractor combining semantic features from an off-the-shelf object instance segmentation network with manually-engineered features such as color histograms, which improves track association when multiple players, usually players from opposing teams, are in physical contact and have nearly coincident boxes. We implement a track re-identification post-processor that connects tracks of players disappearing and reappearing due to camera field of view, also using the track appearance features above. We develop a field pitch element detector based on the baseline pitch semantic segmentation network in the Camera Calibration track, and with its help implement a camera extrinsic parameter estimator to detect drastic camera movements, such as camera panning to follow a long ball, which helps tracking association by factoring out the camera motions from the screen-space track motions.

\mysection{T9 - Movement Forecasting(MF) in Soccer Player Tracking}\\
\textit{Sangrok Lee}\\
\textit{lsrock1@yonsei.ac.kr}

We introduce MF, a novel architecture placed on top of Deep\-LabV3 to perform movement forecasting of players. In SoccerNet tracking dataset, players move dynamically and cameras also move around to follow a ball. Because of that, it is really hard to apply existing tracking methods like Kalman Filter. To tackle this problem, MF is designed to track dynamically moving players under moving camera conditions. MF predicts 2-dimensional movement vectors of players with only one feed-forwarding step and estimates a future position of objects. Although the MF is precise, it is impossible to recognize the person who disappears and reappears. Therefore, we also adopt a re-identification (re-id) module. Overall tracking process consists of the movement estimation and re-id.

\end{document}